\pdfoutput=1

\documentclass[11pt]{article}

\usepackage{soul}%
\usepackage[table]{xcolor} %

\usepackage{ACL2023}

\usepackage{times}
\usepackage{latexsym}

\usepackage[T1]{fontenc}

\usepackage[utf8]{inputenc}

\usepackage{microtype}

\usepackage{inconsolata}

\usepackage{graphicx}
\usepackage{siunitx}
\usepackage{booktabs, multirow}
\usepackage{float}
\usepackage{amssymb}
\usepackage{subcaption}
\usepackage{bm}

\definecolor{Level1}{rgb}{1,0.84,0.7}
\definecolor{Level1Light}{rgb}{1,0.96,0.88}
\definecolor{Level2}{rgb}{0.58,0.87,0.58}
\definecolor{Level2Light}{rgb}{0.95,1,0.92}
\definecolor{Level3}{rgb}{1,1,0.66}
\definecolor{Level3Light}{rgb}{1,1,0.86}
\definecolor{Level4}{rgb}{0.17,0.51,0.83}
\definecolor{Level4Light}{rgb}{0.93,0.99,1}
\definecolor{Gray}{rgb}{0.9,0.9,0.9}
\definecolor{GrayLight}{rgb}{0.97,0.97,0.97}
\definecolor{ForestGreen}{RGB}{34,139,34}

\newcommand{\qualerr}[1]{{\textcolor{red}{#1}}}
\newcommand{\qualfix}[1]{{\textcolor{ForestGreen}{#1}}}

\title{VisText: A Benchmark for Semantically Rich Chart Captioning}

\author{Benny J. Tang$^*$ \\
  MIT CSAIL \\
  \texttt{benjtang@csail.mit.edu} \\\And
  Angie Boggust$^*$ \\
  MIT CSAIL \\
  \texttt{aboggust@csail.mit.edu} \\\And
  Arvind Satyanarayan \\
  MIT CSAIL \\
  \texttt{arvindsatya@mit.edu} \\}

\begin{document}

\maketitle

\def\thefootnote{*}\footnotetext{Both authors contributed equally to this work.}
\def\thefootnote{\arabic{footnote}}

\begin{abstract}
Captions that describe or explain charts help improve recall and comprehension of the depicted data and provide a more accessible medium for people with visual disabilities.
However, current approaches for automatically generating such captions struggle to articulate the perceptual or cognitive features that are the hallmark of charts (e.g., complex trends and patterns).
In response, we introduce VisText: a dataset of 12,441 pairs of charts and captions that describe the charts' construction, report key statistics, and identify perceptual and cognitive phenomena. 
In VisText, a chart is available as three representations: a rasterized image, a backing data table, and a \textit{scene graph}\,---\,a hierarchical representation of a chart's visual elements akin to a web page's Document Object Model (DOM).
To evaluate the impact of VisText, we fine-tune state-of-the-art language models on our chart captioning task and apply prefix-tuning to produce captions that vary the semantic content they convey. 
Our models generate coherent, semantically rich captions and perform on par with state-of-the-art chart captioning models across machine translation and text generation metrics. 
Through qualitative analysis, we identify six broad categories of errors that our models make that can inform future work.

\end{abstract}

\section{Introduction}

Studies have shown that captions can improve the recall and comprehension of the data that charts depict~\citep{hegarty1993constructing, large1995multimedia}.
For instance, when a caption emphasizes visually prominent features of a chart, like a peak or a sharply declining trend, readers treat this information as the key takeaway~\citep{kim2021towards}.
Moreover, for people with visual disabilities, captions (or equivalent descriptions such as alt text) are often the only means of accessing the presented data.
However, as evidenced by numerous guidelines~\citep{jung2021communicating}, producing high-quality chart captions is a non-trivial and laborious manual process.
Thus, despite these advantages, charts are only rarely captioned in practice~\citep{lundgard2022accessible}.

To bridge this gap, several research communities have begun to explore methods for automatically generating chart captions, including using templates and heuristics~\citep{demir2008generating, srinivasan2018augmenting}, adapting image captioning techniques~\citep{balaji2018chart, chen2019neural}, or via data-to-text machine translation~\citep{kantharaj2022chart, obeid2020chart}.
While promising, these approaches have largely produced captions that either describe a chart's construction (e.g., \textit{``The graph is plot between 'Number of people' x-axis over 'Movie Genres' y-axis''}~\citep{balaji2018chart}) or provide statistical summaries (e.g., \textit{``Machinery and equipment was the most valuable commodity for Singapore in 2019''}~\citep{kantharaj2022chart}). 
However, these captions do not articulate the perceptual and cognitive features that make charts a distinctive and compelling medium for communicating data (e.g., \textit{``Prices of Big Tech corporations seem to fluctuate but nevertheless increase over time''}~\citep{lundgard2022accessible}).
Indeed, as \citet{lundgard2022accessible} find, both sighted and blind readers strongly prefer captions that express this type of content. 

To automatically produce such semantically richer captions, we introduce VisText: a benchmark dataset of 12,441 pairs of charts and captions.
VisText makes two key extensions over prior approaches.
First, VisText offers three representations of charts: a rasterized image and backing data table, as in previous work; and a \textit{scene graph}, a hierarchical representation akin to a web page's Document Object Model (DOM), that presents an attractive midpoint between the affordances of chart-as-image and chart-as-data-table. 
Second, for each chart, VisText provides a synthetically generated caption detailing its construction as well as a crowdsourced caption describing its statistical, perceptual, and cognitive features.
These crowdsourced captions represent a substantial increase in data over prior comparable datasets~\citep{mahinpei2022linecap, kantharaj2022chart}.

To demonstrate the possible uses of the VisText dataset, we train three classes of models\,---\,text-based caption models, image-guided captioning models, and semantic prefix-tuning.
Text-based captioning models fine-tune large language models for VisText's chart captioning task, revealing that both data table and scene graph representations can produce compelling and semantically rich captions.
Following recent advancements in image-guided translation~\citep{sulubacak2020multimodal}, we leverage the additional visual affordances in chart images to develop image-guided chart captioning models.
Finally, since users often have varying preferences about the type of semantic content in their captions~\citep{lundgard2022accessible}, we apply semantic prefix-tuning to each of our models, enabling them to output customizable captions.

Our models generate coherent, semantically rich captions across the VisText charts.
Evaluating against standard machine translation and text generation metrics reveals that our models consistently output captions that accurately describe the chart's construction, such as its chart type, title, and axis ranges. 
Through qualitative analysis of our model's captions, we find that our model competently outputs semantically rich captions that describe data trends and complex patterns. 
Further, we categorize six common captioning errors that can inform the future development of chart captioning models on the VisText dataset.

The VisText dataset and source code are available at: \url{https://github.com/mitvis/vistext}.
\section{Related work}

\textbf{Heuristic-Based Chart Captioning.}
Automatically generating natural language descriptions of data tables dates back to \citet{reiter1997building}.
\citet{demir2008generating, demir2010interactive, demir2012summarizing} survey this early work and describe the process of extracting insights from a chart by evaluating a list of propositions and composing selected propositions together to produce a natural language summary.
More recently, data visualization researchers have explored heuristics that calculate summary statistics and templates to assemble natural language ``data facts''~\citep{srinivasan2018augmenting} or descriptions~\citep{cui2019datasite}.
While useful, these approaches yield standardized descriptions that lack the variation and linguistic construction that characterize semantically rich captions~\citep{lundgard2022accessible}.

\textbf{Chart Captioning as Image Captioning.}
With rapid advances of neural image captioning~\citep{vinyals2015show, anderson2017bottom}, researchers have begun to adapt these methods for captioning charts. 
For instance, \citet{balaji2018chart} develop a deep learning pipeline that ingests a PNG chart image, classifies the chart type, detects and classifyies textual content present in the chart, and uses this information to generate a textual description.
\citet{chen2019neural, chen2019figure, chen2020figure} propose a simpler workflow using ResNet to encode the chart image and an LSTM with Attention to decode it into a natural language description. 
Both approaches share a pair of limitations. 
The captions they produce convey relatively simplistic information about the chart (e.g., title, axis labels, etc.) or articulate concepts in visual rather than data terms (e.g., \textit{``Dark Magenta has the lowest value''}).
While both approaches contribute associated datasets, their charts and captions are synthetically generated and may not represent real-world counterparts. 
SciCap~\citep{hsu-etal-2021-scicap-generating} addresses this limitation by scraping real-world charts from 290,000 arXiv papers; however, the baseline models trained on this dataset struggle to generate semantically rich captions.

\textbf{Chart Captioning as Text Translation.}
Perhaps closest to our contribution is recent work modeling chart captioning as a data-to-text problem.
For instance, \citet{spreafico202} train an encoder-decoder LSTM architecture to generate a natural language caption from time series data. 
Similarly, \citet{obeid2020chart} and \citet{kantharaj2022chart} explore how transformer architectures can translate tabular structures into captions. 
These data-to-text methods are more successful than chart-as-image captioning, yielding captions that better capture relevant information from the charts and have higher BLEU scores. 
Nevertheless, we observe two limitations with these data-to-text approaches that motivate our contribution. 
First, data-to-text methods are heavily reliant on access to a chart's data table.
In practice, data tables are only rarely published alongside charts and methods that recover equivalent information via OCR experience a significant drop in performance~\citep{kantharaj2022chart}.
Second, the associated datasets do not contain sufficient training examples of captions that express semantically rich insights about the depicted data (i.e., the perceptual and cognitive phenoma that distinguish charts as a medium as distinct from data tables~\citep{lundgard2022accessible}).
As a result, while the generated captions are compelling, they are largely limited to reporting statistics which sighted and blind readers prefer less than captions that convey complex trends and patterns~\citep{lundgard2022accessible}.

\section{The VisText Dataset}

\begin{figure*}[t]
  \centering
  \includegraphics[width=\textwidth]{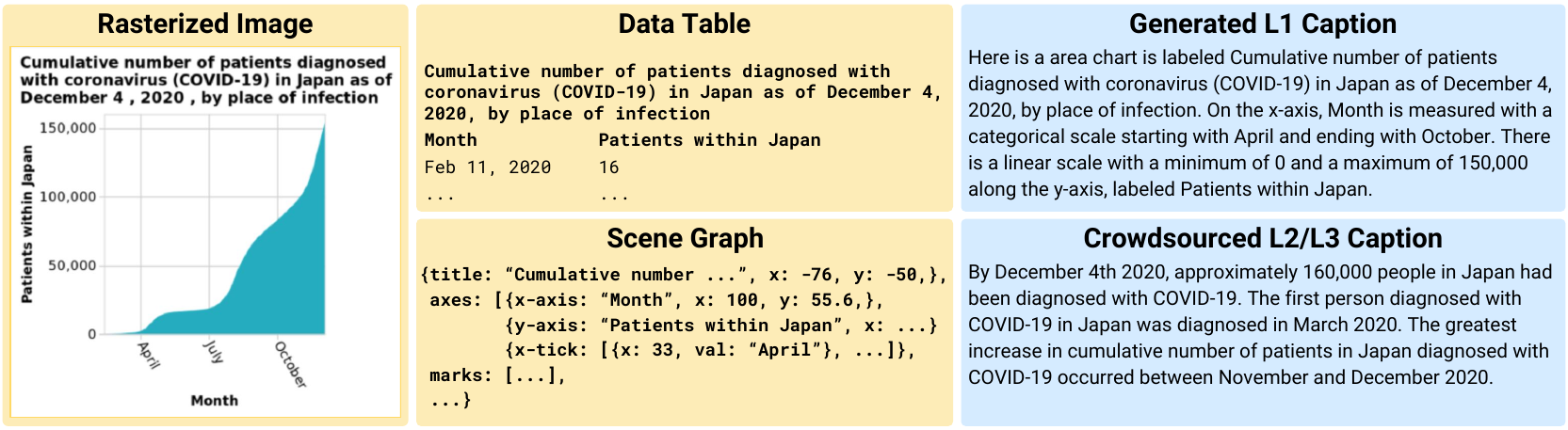}
  \caption{The VisText dataset consists of 12,441 charts represented as a rasterized image, data table, and scene graph. Before model training, each data table and scene graph is processed from its original form (shown) to a minimized and linearized text representation. Each chart is accompanied by a generated L1 caption describing the aspects of the chart's construction (e.g., chart type and axis labels) and a crowdsourced L2/L3 caption describing summary statistics and interesting trends~\cite{lundgard2022accessible}.}
  \vspace{-5mm}
  \label{fig:dataset}
\end{figure*}

We designed the VisText dataset in response to two limitations existing datasets present for generating semantically rich chart captions. First, existing datasets represent charts as either rasterized images or as data tables.
While useful, these representations trade off perceptual fidelity and chart semantics in mutually exclusive ways\,---\,images capture the perceptual and cognitive phenomena that are distinctive to charts (e.g., trends or outliers) but pixels cannot express the rich semantic relationships between chart elements (e.g., estimating plotted data values using axis labels).
While the vice-versa is true~\citep{lundgard2022accessible}, tables also present additional caveats. 
There is not always a one-to-one relationship between the semantics of a data table and chart (i.e., one data table may be the source for several distinctly different charts).
Moreover, data tables are rarely published alongside charts; and, automatic data table extraction is error-prone due to the diversity of chart types and visual styles as well as the difficulty of reasoning about visual occlusion~\citep{kantharaj2022chart, luo2021chartocr, jung2017chartsense}).

Second, if existing datasets provide captions that describe perceptual or cognitive features, these captions comprise only a small portion of the dataset.
At best, LineCap~\citep{mahinpei2022linecap} offers 3,528 such captions for line charts only, while Chart-to-Text~\citep{kantharaj2022chart} estimates that roughly 15\% of the sentences in its captions across a variety of chart types express such content.

In contrast, VisText provides 12,441 crowdsourced English captions that articulate statistical, perceptual, and cognitive characteristics of bar, line, and area charts. 
In VisText, charts are available as not only data tables and rasterized images but also as \textit{scene graphs}.
Scene graphs are hierarchical representations that better preserve perceptual fidelity and chart semantics, are often the format for publishing web-based charts, and can be recovered from chart images~\citep{poco2017reverse}.

\subsection{Data Table Collection}

The data tables found in VisText are sourced from the Statista dataset of the Chart-to-Text benchmark~\citep{kantharaj2022chart}.
The tables were collected by crawling Statista.com in December 2020 and contain real-world data related to technology, trade, retail, and sports.
We process these tables to make them amenable for chart generation, including stripping formatting symbols (e.g., \texttt{\$} and \texttt{\%}), standardizing data strings, and identifying the measure type of each column (i.e., quantitative, categorical, or temporal). 
Data tables are discarded if they do not contain at least one quantitative field and one categorical or temporal field, or if other errors occur during the processing steps.
We further down select to data tables containing between 2 to 20 columns and 10 to 500 rows.
If a data table has over 500 rows, we randomly sample rows.
In larger data tables, this step potentially affects how salient a trend is.

\subsection{Chart Generation and Representation}
\label{sec:chart_gen}

Charts in the Chart-to-Text Statista dataset all feature the same layout and visual appearance.
In contrast, we aim for richer visual diversity by generating charts using the Vega-Lite visualization library~\citep{satyanarayan2016vegalite} via the Python Altair package~\citep{vanderplas2018altair}.
To facilitate collecting high-quality captions, we focus on univariate charts: charts that depict one quantitative observation against a categorical or temporal variable. 
This focus is informed by recent work in the data visualization research community which has chosen single-series line charts as the target of study for natural language descriptions~\citep{kim2021towards,stokes2022striking}.
VisText also includes single-series bar and area charts as they typically exhibit similar perceptual features to line charts.

For each data table, we iterate through pairs of univariate fields. 
If the pair contains a temporal field, we randomly generate an area or line chart; if the pair contains a categorical field, we randomly generate a horizontal or vertical bar chart. 
For diversity in layout and visual appearance, we randomly rotate axis labels 
and apply one of fourteen themes provided by the Vega-Lite library. 
These themes mimic the visual style of common chart platforms or publishers (e.g., ggplot2 or the LA Times).

In VisText, each chart is represented as a \textit{rasterized image}, stored as an RGBA-encoded PNG file, as well as a \textit{scene graph}. 
A scene graph is a textual representation of the rendered chart similar to a web page's Document Object Model (DOM). 
Scene graphs encode the position, value or content, and semantic role of all visual elements within a chart, including the individual marks (i.e., bars or points along the line), titles, axes gridlines, etc.
Thus, scene graphs express the perceptual features of rasterized images in a more computationally-tractable form.

Scene graphs are a standard data structure for representing vector-based graphics\,---\,the most common format for publishing visualizations\,---\,and, thus, can be trivially recovered (e.g., by traversing the SVG text string).
We extract the scene graph directly from the rendered chart using the Vega-Lite API.
As most text generation models expect a linear set of input tokens, we flatten the scene graph via a depth-first traversal.
To scale to large language models, we need to further reduce the size of the scene graph.
Thus, we preserve the following elements which we hypothesize as being most critical for generating semantically rich captions: \texttt{title}, \texttt{title coordinates}, \texttt{axis labels}, \texttt{axis label coordinates}, \texttt{axis tick coordinates}, \texttt{mark coordinates}, and \texttt{mark sizes}.
VisText includes both the original (hierarchical) and reduced (linearized) scene graphs.

\subsection{Caption Generation and Collection}

\begin{figure*}[t]
  \centering
  \includegraphics[width=\textwidth]{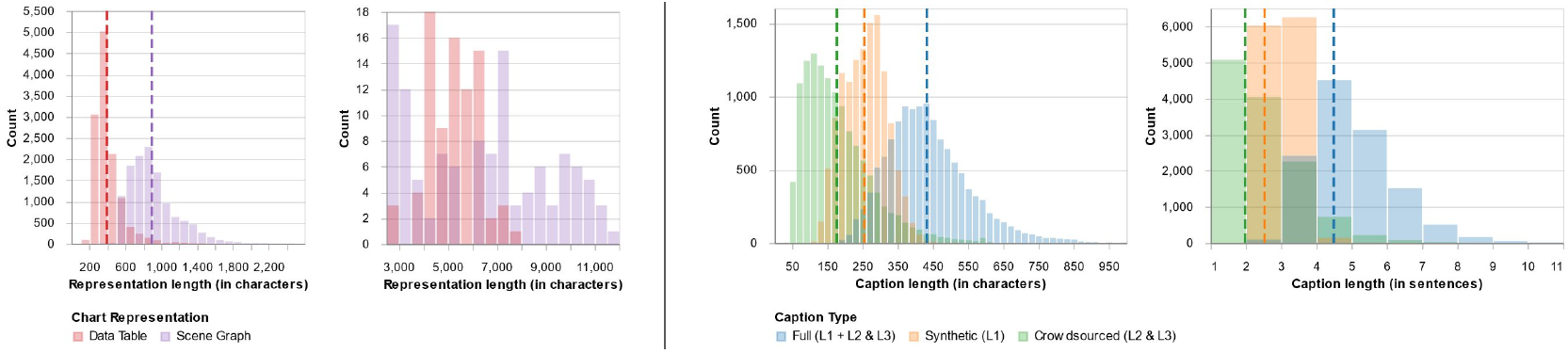}
    \caption{The VisText dataset contains data table and scene graph representations of each chart paired with L1 and L2/L3 captions. The distributions and means (dotted lines) of representations (left pair) and captions (right pair) are shown. As the distribution of chart representations has a long tail, we split it into two charts at 2,500 characters to better display the tail by re-scaling the y-axis of the second chart.}
  \label{fig:dataset_profile}
  \vspace{-5mm}
\end{figure*}

Our captioning process is guided by the framework developed by~\citet{lundgard2022accessible}, which identifies four levels of semantic content: \textbf{L1} content enumerates aspects of the chart's construction (e.g., axis ranges); \textbf{L2} content reports summary statistics and relations (e.g., extrema); \textbf{L3} content synthesizes perceptual and cognitive phenomena (e.g., complex trends); and, \textbf{L4} content describes domain-specific insights (e.g., sociopolitical context).
In subsequent studies, the authors find that while sighted readers typically prefer higher levels of semantic content, blind readers are split about the usefulness of L1 and L4 content. 
Thus, given these differing preferences, we define a single caption to express multiple levels of content separated across clauses or sentences. 
We only consider the first three levels of this model, and leave L4 content to future work.
Following guidelines prescribed by the National Center for Accessible Media (NCAM), our captions begin with L1 content and then turn to L2 and L3 content~\citep{gould2008effective}.

We algorithmically generate L1 content and use a crowdsourced protocol to collect L2 and L3 content. 
This approach follows ~\citep{lundgard2022accessible}'s computational considerations as well as results from \citet{morash2015guiding} who find that, even with instructions and guidelines, crowd workers do not describe a chart's structural elements sufficiently for blind readers.
Thus, synthetically generating L1 content allows us to ensure that captions convey complete descriptions of the chart's structural elements.
L1 content comprises 1 sentence conveying the chart type and title, and then 1\,--\,2 sentences describing the axes (including the titles, ranges, and scales).
We use template randomization to generate a diverse range of L1 captions to mimic human variability and reduce the capacity of the model to overfit to a single L1 style.
Three templates are defined for the first sentence and twenty-six template combinations for the subsequent sentences.
During generation, we randomly select a pair of templates and fill in information from the abstract chart specification.
For additional diversity, we randomly drop scale information and swap template words with synonyms. 
Templates and synonym replacements are listed in Appendix~\ref{sec:l1_generation}.

To crowdsource L2 and L3 content, we extend the protocol used by~\citet{lundgard2022accessible}. 
After soliciting consent, we introduce the task: participants are presented with a chart image and corresponding L1 description; they are asked to write a description about the trends and patterns they observe without drawing on background knowledge or repeating L1 information.
The introduction provides examples and explanations of valid and invalid responses.
After acknowledging these examples, participants are asked to complete 5 random iterations of the task. 
To maximize the quality of our crowdsourced captions, we manually curated the charts and L1 descriptions used in the study. 
We discarded any charts that were challenging to read (e.g., colors were too similar, marks were not easily readable, etc.). 
Participants were recruited on the Prolific.co platform, took approximately 14 minutes to complete the study, and were compensated \$3.25 (\$14/hour).
Additional details on our crowdsourcing process are in Appendix~\ref{sec:crowdsource_materials}.

We manually verified charts where participants failed an attention check and discarded invalid descriptions.
Additionally, we manually inspected captions for personally identifiable information or offensive content.
Using heuristics, we removed captions where respondents described charts as unclear or illegible and replaced newline characters with spaces.
Although we attempted to fix incorrect spelling and casing errors using a similar heuristic-based approach, we observed that this process could improperly affect axis and chart names.
As a result, these errors remain in our dataset.

\subsection{Dataset Analysis}

Figure~\ref{fig:dataset_profile} shows the distribution and means of the lengths of chart representations and captions.
Synthetically generated L1 captions have roughly 1.5x more characters than crowdsourced L2/L3 captions ($\mu=255$ vs. $\mu=177$) but the average number of sentences are comparable (2.5 vs. 2).
The VisText dataset consists of captions for 3,189 area charts, 6,238 bar charts, and 3,014 line charts\,---\,the roughly twice-as-many bar charts as area or line charts corresponds to the randomization of temporal fields during chart generation (Sec.~\ref{sec:chart_gen}).
As some charts have multiple crowdsourced captions, we randomly split our dataset into training, validation, and test sets using the chart IDs to prevent data leakage across sets.
This resulted in an approximate ratio of 80:10:10.

Finally, to understand the distribution of semantic content, we manually coded 2\% (230) of crowdsourced captions. 
We followed a protocol inspired by \citet{lundgard2022accessible} by breaking sentences down into independent statements and mapping these statements to their semantic content level.
We marked statements as not categorizable if they did not map to the framework\,---\,for instance, if captions expressed commentary from crowd workers such as \textit{``this chart is hard to read.''}
Our analysis revealed 11 L1 statements (2.4\%), 180 L2 statements (39.7\%), 253 L3 statments (55.7\%), and 10 not categorizable statements (2.2\%).
While a handful express L1 content, the bulk of statements (95\%) express L2 or L3 content, with approximately 1.4x L3 statements than L2.

\section{Chart Captioning Models}
\label{sec:models}

To demonstrate the affordances of the VisText dataset, we train three classes of models. 
First, we fine-tune large language models to translate from textual chart representations to natural language captions. 
These models evaluate the feasibility and impact of \texttt{scene-graph} models compared to prior \texttt{data-table} approaches~\citep{kantharaj2022chart}. 
Second, as VisText provides multiple chart representations, we adapt image-guided translation~\citep{sulubacak2020multimodal, cho2021unifying} to develop two multimodal chart captioning models: \texttt{image-scene-graph} and \texttt{image-data-table}.
Finally, since VisText offers captions at different semantic levels and prior work has shown significant differences in readers' preferences~\citep{lundgard2022accessible}, we explore prefix-tuned models that selectively output L1, L2/L3, or L1+L2/L3 captions.
Training details are in Appendix~\ref{sup:trainingdetails}.

\subsection{Text-Based Chart Captioning}

Informed by prior work~\citep{kantharaj2022chart}, we investigate text translation models for generating chart captions.
In particular, Kantharaj et al. found that models that translate data tables to chart captions significantly outperform image captioning models. 
However, when data tables were not available, the authors found a significant drop in their models' ability to extract relevant information from the chart\,---\,an effect that was only slightly ameliorated by using OCR methods to extract text from chart images.
In contrast, VisText's scene graphs can be more readily recovered from charts when data tables are not available\,---\,for instance, by processing the SVG format of web-based visualizations. 
Moreover, scene graphs offer a potentially richer source of information than data tables as they encode visual properties of the chart (e.g., coordinates and colors) and are less noisy than tokens recovered via OCR. 
Thus, to evaluate the feasibility and efficacy of scene graphs, we train a \texttt{scene-graph} text translation model and a baseline \texttt{data-table} model for comparison.

For each model, we fine-tune a pretrained ByT5 transformer model~\citep{byt5} on the VisText dataset.
We choose ByT5 over T5 transformers~\citep{T5} because it uses a token-free, byte-encoding that eliminates the use of a tokenizer. 
As a result, it is robust to noisy inputs, minimizes the need for text preprocessing, and eliminates the out-of-dictionary problem. 
This allows our model to handle common typographical and chart reading errors in the crowdsourced L2 and L3 captions and increases generalizability to previously-unseen words that could be present in chart and axes titles.

\subsection{Image-Guided Chart Captioning}
Following recent advancements in image-guided machine translation~\citep{sulubacak2020multimodal}, we train image-guided captioning models using the VisText dataset. 
Images have improved text-based machine translation models by providing visual information complementary to natural language inputs. 
Similarly, chart images can contain visuals complementary to the textual specification. 
For instance, visual affordances that are important for perceiving a trend (e.g., gestalt relations, relative sizes/areas, etc.) may be obfuscated in the scene graph but better captured in the chart image.

We train three image-guided chart captioning models: \texttt{image}, \texttt{image-scene-graph}, and \texttt{image-data-table}. 
All models leverage the vision-language transformer model VL-T5~\citep{cho2021unifying}. 
VL-T5 is pretrained on image captioning and visual grounding tasks and was successfully applied to machine translation, making it suitable for chart captioning. 
We extract visual features for each VisText chart image using a Bottom-Up Feature Extractor~\citep{anderson2017bottom}.
To explore the value of images to chart captioning, our \texttt{image} model only takes in the image features, while \texttt{image-scene-graph} and \texttt{image-data-table} concatenate the image features with the chart's textual representations (scene graph or data table).

\subsection{Semantic Prefix-Tuning}
In real-world chart captioning settings, users want to vary the level of semantic content in their captions. 
For instance, while some blind users want verbose captions that describe the chart visuals, sighted users may only want captions that help them expose data trends~\citep{lundgard2022accessible}. 
To develop models capable of such customization, we leverage prefix-tuning strategies alongside VisText's semantic caption breakdown.
Prefix-tuning specifies a task alongside the input, permitting a single large language model to perform many different tasks. 
In our setting, we use prefix-tuning to specify the level of semantic content to include in the caption~\citep{prefixtuning}.

We train each of our models with and without semantic prefix-tuning.
With semantic prefix-tuning, we treat chart captioning as a multi-task fine-tuning problem, where the model is trained to generate the L1 and L2/L3 captions separately. 
In every epoch, the model sees each VisText chart twice, once with the L1 prefix and caption and once with the L2/L3 prefix and caption.

\section{Evaluation and Results}

To evaluate the VisText dataset and our chart captioning models, we measure the readability and accuracy of generated captions and their similarity to the VisText target caption. 
We also qualitatively analyze the descriptiveness of generated L2/L3 captions and categorize common errors.

\subsection{Quantitative Model Performance}
\label{sec:quanteval}
\begin{table*}
\centering
\resizebox{\linewidth}{!}{
\begin{tabular}{lrrrrrrrrrrr}\toprule
\textbf{Input} &\textbf{PT} &\textbf{BLEU $\uparrow$} &\textbf{Perplexity $\downarrow$} &\textbf{RG $\uparrow$} &\textbf{ROUGE-1 $\uparrow$} &\textbf{ROUGE-2 $\uparrow$} &\textbf{ROUGE-L $\uparrow$} &\textbf{ROUGE-L SUM $\uparrow$} &\textbf{WMD $\downarrow$} &\textbf{TER $\downarrow$} \\\midrule
\citet{kantharaj2022chart} & &\cellcolor[HTML]{e1eff8}$0.30 \pm 1.27\mathrm{e}{-3}$ &\cellcolor[HTML]{f7fbff}$28.51 \pm 1.02\mathrm{e}{-1}$ &\cellcolor[HTML]{a8d0e5}$1.69 \pm 8.13\mathrm{e}{-3}$ &\cellcolor[HTML]{b1d5e8}$0.58 \pm 8.67\mathrm{e}{-4}$ &\cellcolor[HTML]{bfdcec}$0.42 \pm 1.73\mathrm{e}{-3}$ &\cellcolor[HTML]{add3e7}$0.49 \pm 9.33\mathrm{e}{-4}$ &\cellcolor[HTML]{afd4e7}$0.49 \pm 9.67\mathrm{e}{-4}$ &\cellcolor[HTML]{c1dded}$0.67 \pm 2.43\mathrm{e}{-3}$ &\cellcolor[HTML]{9fcae1}$66.99 \pm 4.88\mathrm{e}{-2}$ \\
\citet{kantharaj2022chart} &\checkmark &\cellcolor[HTML]{f3f9fe}$0.30 \pm 2.23\mathrm{e}{-3}$ &\cellcolor[HTML]{f7fbff}$31.15 \pm 9.73\mathrm{e}{-1}$ &\cellcolor[HTML]{a8d0e5}$1.69 \pm 8.67\mathrm{e}{-4}$ &\cellcolor[HTML]{abd1e6}$0.59 \pm 1.20\mathrm{e}{-3}$ &\cellcolor[HTML]{b2d5e8}$0.43 \pm 1.47\mathrm{e}{-3}$ &\cellcolor[HTML]{a4cee3}$0.49 \pm 1.60\mathrm{e}{-3}$ &\cellcolor[HTML]{a4cee3}$0.49 \pm 1.60\mathrm{e}{-3}$ &\cellcolor[HTML]{bbdaea}$0.67 \pm 2.33\mathrm{e}{-3}$ &\cellcolor[HTML]{9ecae1}$66.97 \pm 3.07\mathrm{e}{-1}$ \\\midrule
\texttt{scene-graph} & &\cellcolor[HTML]{a8cfe5}$0.32 \pm 4.07\mathrm{e}{-3}$ &\cellcolor[HTML]{c3deed}$20.96 \pm 3.09\mathrm{e}{+0}$ &\cellcolor[HTML]{9ecae1}$1.82 \pm 2.67\mathrm{e}{-4}$ &\cellcolor[HTML]{d1e6f2}$0.56 \pm 1.42\mathrm{e}{-2}$ &\cellcolor[HTML]{e4f1f9}$0.39 \pm 1.62\mathrm{e}{-2}$ &\cellcolor[HTML]{cbe3f1}$0.47 \pm 1.04\mathrm{e}{-2}$ &\cellcolor[HTML]{cbe3f1}$0.47 \pm 1.04\mathrm{e}{-2}$ &\cellcolor[HTML]{d8eaf4}$0.68 \pm 8.33\mathrm{e}{-3}$ &\cellcolor[HTML]{add2e6}$69.34 \pm 2.31\mathrm{e}{+0}$ \\
\texttt{data-table} & &\cellcolor[HTML]{a5cee4}$0.32 \pm 2.40\mathrm{e}{-3}$ &\cellcolor[HTML]{c1ddec}$20.65 \pm 2.15\mathrm{e}{+0}$ &\cellcolor[HTML]{a8d0e5}$1.69 \pm 1.27\mathrm{e}{-3}$ &\cellcolor[HTML]{cfe5f2}$0.56 \pm 7.30\mathrm{e}{-3}$ &\cellcolor[HTML]{e3f0f8}$0.39 \pm 8.83\mathrm{e}{-3}$ &\cellcolor[HTML]{cbe3f1}$0.47 \pm 6.20\mathrm{e}{-3}$ &\cellcolor[HTML]{cde4f1}$0.47 \pm 6.13\mathrm{e}{-3}$ &\cellcolor[HTML]{ecf5fb}$0.68 \pm 3.60\mathrm{e}{-3}$ &\cellcolor[HTML]{b2d5e7}$70.21 \pm 7.90\mathrm{e}{-1}$ \\
\texttt{scene-graph} &\checkmark &\cellcolor[HTML]{a5cee4}$0.32 \pm 2.13\mathrm{e}{-3}$ &\cellcolor[HTML]{bddbeb}$20.02 \pm 2.25\mathrm{e}{+0}$ &\cellcolor[HTML]{a2cce3}$1.78 \pm 4.25\mathrm{e}{-2}$ &\cellcolor[HTML]{d1e6f2}$0.56 \pm 6.70\mathrm{e}{-3}$ &\cellcolor[HTML]{e4f1f9}$0.39 \pm 6.23\mathrm{e}{-3}$ &\cellcolor[HTML]{cfe5f2}$0.47 \pm 6.37\mathrm{e}{-3}$ &\cellcolor[HTML]{cfe5f2}$0.47 \pm 6.40\mathrm{e}{-3}$ &\cellcolor[HTML]{dfeef7}$0.68 \pm 1.23\mathrm{e}{-2}$ &\cellcolor[HTML]{c0ddec}$72.55 \pm 1.75\mathrm{e}{+0} $\\
\texttt{data-table} &\checkmark &\cellcolor[HTML]{aed3e7}$0.32 \pm 4.23\mathrm{e}{-3}$ &\cellcolor[HTML]{daebf5}$24.23 \pm 1.81\mathrm{e}{+0}$ &\cellcolor[HTML]{a5cee4}$1.73 \pm 8.65\mathrm{e}{-2}$ &\cellcolor[HTML]{c1dded}$0.57 \pm 5.90\mathrm{e}{-3}$ &\cellcolor[HTML]{d4e8f3}$0.40 \pm 5.57\mathrm{e}{-3}$ &\cellcolor[HTML]{c2deed}$0.48 \pm 5.53\mathrm{e}{-3}$ &\cellcolor[HTML]{c2deed}$0.48 \pm 5.60\mathrm{e}{-3}$ &\cellcolor[HTML]{c8e1ef}$0.67 \pm 1.63\mathrm{e}{-3}$ &\cellcolor[HTML]{b2d5e8}$70.29 \pm 2.04\mathrm{e}{+0}$ \\ \midrule
\texttt{image} & &\cellcolor[HTML]{f7fbff}$0.07 \pm 1.07\mathrm{e}{-3}$ &\cellcolor[HTML]{aad1e5}$17.36 \pm 9.46\mathrm{e}{-1}$ &\cellcolor[HTML]{e9f4fb}$0.78 \pm 1.04\mathrm{e}{-2}$ &\cellcolor[HTML]{f7fbff}$0.34 \pm 5.87\mathrm{e}{-3}$ &\cellcolor[HTML]{f7fbff}$0.14 \pm 3.60\mathrm{e}{-3}$ &\cellcolor[HTML]{f7fbff}$0.25 \pm 4.03\mathrm{e}{-3}$ &\cellcolor[HTML]{f7fbff}$0.25 \pm 4.07\mathrm{e}{-3}$ &\cellcolor[HTML]{f7fbff}$1.11 \pm 7.10\mathrm{e}{-3}$ &\cellcolor[HTML]{f7fbff}$89.03 \pm 9.12\mathrm{e}{-1}$ \\
\texttt{image-scene-graph} & &\cellcolor[HTML]{f0f7fd}$0.30 \pm 3.83\mathrm{e}{-3} $ &\cellcolor[HTML]{f5f9fe}$28.15 \pm 2.26\mathrm{e}{+0} $ &\cellcolor[HTML]{9fcbe2}$1.82 \pm 2.50\mathrm{e}{-3} $ &\cellcolor[HTML]{a6cfe4}$0.59 \pm 1.20\mathrm{e}{-3} $ &\cellcolor[HTML]{a4cee3}$0.43 \pm 2.47\mathrm{e}{-3} $ &\cellcolor[HTML]{aad1e5}$0.49 \pm 2.53\mathrm{e}{-3} $ &\cellcolor[HTML]{aad1e5}$0.49 \pm 2.53\mathrm{e}{-3} $ &\cellcolor[HTML]{b1d4e7}$0.66 \pm 1.53\mathrm{e}{-3} $ &\cellcolor[HTML]{a1cce2}$67.45 \pm 2.82\mathrm{e}{-1}$ \\
\texttt{image-data-table} & &\cellcolor[HTML]{f7fbff}$0.29 \pm 1.20\mathrm{e}{-3}$ &\cellcolor[HTML]{f7fbff}$29.81 \pm 2.62\mathrm{e}{-1}$ &\cellcolor[HTML]{9fcbe2}$1.81 \pm 1.20\mathrm{e}{-3}$ &\cellcolor[HTML]{9ecae1}$0.59 \pm 5.67\mathrm{e}{-4}$ &\cellcolor[HTML]{9ecae1}$0.44 \pm 1.03\mathrm{e}{-3}$ &\cellcolor[HTML]{9ecae1}$0.49 \pm 2.17\mathrm{e}{-3}$ &\cellcolor[HTML]{9ecae1}$0.49 \pm 2.23\mathrm{e}{-3}$ &\cellcolor[HTML]{9ecae1}$0.66 \pm 6.33\mathrm{e}{-4}$ &\cellcolor[HTML]{9ecae1}$66.80 \pm 2.77\mathrm{e}{-2}$\\
\texttt{image} &\checkmark &\cellcolor[HTML]{f7fbff}$0.07 \pm 1.33\mathrm{e}{-3}$ &\cellcolor[HTML]{d9eaf4}$24.08 \pm 1.77\mathrm{e}{+0}$ &\cellcolor[HTML]{f7fbff}$0.58 \pm 1.34\mathrm{e}{-2}$ &\cellcolor[HTML]{f7fbff}$0.33 \pm 6.20\mathrm{e}{-3}$ &\cellcolor[HTML]{f7fbff}$0.13 \pm 3.17\mathrm{e}{-3}$ &\cellcolor[HTML]{f7fbff}$0.23 \pm 4.67\mathrm{e}{-3}$ &\cellcolor[HTML]{f7fbff}$0.23 \pm 4.67\mathrm{e}{-3}$ &\cellcolor[HTML]{f7fbff}$1.11 \pm 1.90\mathrm{e}{-3}$ &\cellcolor[HTML]{f7fbff}$100.04 \pm 6.57\mathrm{e}{+0} $\\
\texttt{image-scene-graph} &\checkmark &\cellcolor[HTML]{abd1e6}$0.32 \pm 9.90\mathrm{e}{-3}$ &\cellcolor[HTML]{9ecae1}$15.50 \pm 4.45\mathrm{e}{-1}$ &\cellcolor[HTML]{9fcbe2}$1.82 \pm 2.67\mathrm{e}{-4}$ &\cellcolor[HTML]{f7fbff}$0.54 \pm 8.23\mathrm{e}{-3}$ &\cellcolor[HTML]{f7fbff}$0.38 \pm 4.93\mathrm{e}{-3}$ &\cellcolor[HTML]{f7fbff}$0.45 \pm 3.63\mathrm{e}{-3}$ &\cellcolor[HTML]{f7fbff}$0.45 \pm 3.57\mathrm{e}{-3}$ &\cellcolor[HTML]{f7fbff}$0.69 \pm 7.13\mathrm{e}{-3}$ &\cellcolor[HTML]{f7fbff}$81.95 \pm 4.53\mathrm{e}{+0} $\\
\texttt{image-data-table} &\checkmark &\cellcolor[HTML]{9ecae1}$0.32 \pm 2.87\mathrm{e}{-3}$ &\cellcolor[HTML]{aad0e5}$17.29 \pm 1.28\mathrm{e}{+0}$ &\cellcolor[HTML]{9fcbe2}$1.81 \pm 4.50\mathrm{e}{-3}$ &\cellcolor[HTML]{f0f7fd}$0.54 \pm 6.67\mathrm{e}{-3}$ &\cellcolor[HTML]{f3f9fe}$0.38 \pm 7.50\mathrm{e}{-3}$ &\cellcolor[HTML]{f0f7fd}$0.45 \pm 5.60\mathrm{e}{-3}$ &\cellcolor[HTML]{f0f7fd}$0.45 \pm 5.50\mathrm{e}{-3}$ &\cellcolor[HTML]{e2eff8}$0.68 \pm 1.33\mathrm{e}{-3}$ &\cellcolor[HTML]{eef6fc}$80.21 \pm 1.34\mathrm{e}{+0} $\\
\bottomrule
\end{tabular}
}
\caption{We compare our text-based models (\texttt{scene-graph} and \texttt{data-table}), our image-guided models (\texttt{image}, \texttt{image-scene-graph}, and \texttt{image-data-table}), and semantic prefix-tuning (PT) models to prior chart captioning models~\citep{kantharaj2022chart}. We evaluate each model using machine translation and text generation metrics, including BLEU~\citep{bleu}, Perplexity, Relation Generation (RG)~\citep{relationgeneration}, ROUGE~\citep{rouge}, Word Mover's Distance (WMD)~\citep{wmd}, and Translational Error Rate (TER)~\citep{ter1}. We report the mean and standard deviation of three independent models. Darker colors indicate better scores.}
\label{tab:quant-results-table}
\end{table*}

We evaluate the results of our text-based and image-guided captioning models with and without prefix-tuning. 
We also compare to a current state-of-the-art chart captioning model that uses data table chart representations and a T5 generation model~\citep{kantharaj2022chart}. 
To measure the quality of output captions, we evaluate each model on machine translation and language generation metrics (Table~\ref{tab:quant-results-table}).

\paragraph{Chart images do not support captioning.}
The \texttt{image} model performs the worst of all the chart captioning models.
Its low perplexity and high error rates indicate it is highly confident in its inaccurate captions.
While chart images contain the same information encoded in the chart's textual representations, it is presumably not adequately extracted by the model.
Both the image model backbone~\citep{cho2021unifying} and the visual feature extractor~\citep{anderson2017bottom} are trained on natural images, making chart images out-of-distribution inputs that are likely to be poorly represented by these vision models.
As the chart captioning task grows, model backbones, architectures, and feature extractors could be customized to chart images, which may improve image-based chart captioning.

\paragraph{All models produce high quality L1 captions.}
In our chart captioning setting, relation generation~\citep{relationgeneration} measures how often the chart title, axis names, and axis scales in the input appear in the caption. 
Every model (except \texttt{image}) achieves a similarly-high relation generation score, indicating that every model can generate detailed L1 captions.

\paragraph{Scene graphs perform as well as data tables.} 
Models trained on scene graph representations achieve similar performance across the evaluative metrics to models trained on data tables.
As scene graphs can be more easily extracted from web-based charts images, they may be the preferred representation for future chart captioning models.

\paragraph{Image-guiding does not improve captioning.}
Our image-guided captioning models do not experience the significant increase in performance other image-guided translation tasks report.
While in image-guided translation, images contain substantial additional information beyond the text, the image and textual representations in chart captioning often contain highly similar information.
The small amount of additional information in images might benefit complex captioning tasks on multivariate charts or infographics; however, the current VisText captions rarely reference visual information not present in the scene graph or data table.

\paragraph{Prefix-tuning is free.}
Adding semantic prefix-tuning to our models does not significantly change their performance.
Models trained with and without prefix-tuning are exposed to the same set of charts, so it is consistent that prefix-tuning would not impact the quality of output captions.
Given prefix-tuned models are able to output L1, L2/L3, and L1+L2/L3 captions, prefix-tuning may be preferred if users require semantic customization.

\subsection{Qualitative Caption Evaluation}
\label{sec:qualeval}
To augment our quantitative evaluation, we qualitatively assess the descriptiveness and accuracy of the generated chart captions. 
Since L1 caption accuracy can be measured at scale via relation generation, we focus our evaluation on L2/L3 predictions. 

Prior analysis tasked annotators with comparing the accuracy, coherence, and fluency of generated captions compared to a target caption~\citep{kantharaj2022chart}.
Instead, our approach follows an inductive qualitative data analysis approach: iteratively analyzing captions in a ``bottom-up'' fashion to identify emergent patterns in how generated captions compare to the ground truth~\citep{bingham2021deductive}.
We randomly sample 176 generated captions from the \texttt{scene-graph} model with prefix-tuning and break them into their independent L2 and L3 statements, resulting in 181 (48.27\%) L2 statements and 194 (51.73\%) L3 statements.

Approximately half (241 / 512) of the L2 and L3 statements made in the generated captions are factually accurate.
Moreover, many of the full sentences are written in a natural, human-like manner and generated captions frequently include both compound and complex sentences.
On average, every generated caption has one L3 statement and zero to two L2 statements. 
Often this takes the form of a L3 general trend statement (e.g., \textit{``The median annual family income in Canada has increased from 2000 to 2018''}) accompanied by an L2 minimum and maximum statement (\textit{``The highest was in 2015 at 80k and the lowest was in 2000''}).
For the remaining half of analyzed captions, we identified the following recurring types of errors:

\paragraph{Identity Errors.}
We identify 86 identity errors (22.93\% of analyzed statements).
An identity error occurs when an L2 or L3 statement incorrectly reports the independent variable for a given (often correctly identified) trend.
For bar charts, this error means incorrectly reporting the categorical label associated with a bar (e.g., in Appendix Figure \ref{app:qidentity}: \textit{``The most popular music activity is \qualerr{vinyl albums and vinyl singles}''} should be \textit{``The most popular music activity is \qualfix{tickets for festivals}''}).
For area and line charts, this error means incorrectly identifying the temporal point or range of the trend. 
With bar charts, in particular, we observed that the identities were often ``off-by-one'' (i.e., identifying a minimum or maximum value, but attributing it to the second-highest or second-lowest category).

\paragraph{Value Errors.}
A value error occurs when the quantitative data value of a statement is incorrect.
Of the captions we analyzed, 3.20\% (12) of statements contained a value error.
For instance, as shown in Appendix Figure \ref{app:qvalue}, for the caption \textit{``The total gate revenue from sporting events worldwide by region from 2006 to 2015 has increased from around 15 billion dollars to \qualerr{around 15 billion dollars}''}, the value should be \qualfix{around 18 billion dollars}.
If it is ambiguous whether an error is an Identity or Value Error, we classify it as the former.

\paragraph{Direction Errors.}
A direction error occurs when the direction (which can be \textit{increasing}, \textit{decreasing}, or \textit{stable}) of a trend in an L3 statement is incorrect.
We uncovered 32 direction errors (8.53\% of analyzed statements).
For instance, in the caption \textit{``The per capita consumption of sweet corn in the US has \qualerr{increased} from 2000 to 2019''} (Appendix Figure \ref{app:qdirection}), the trend is actually \qualfix{decreased}.
In most direction errors, the identity (i.e., temporal range) is correct.

\paragraph{Stability Errors.}
A stability error occurs when the magnitude of a direction or the variance in a trend is incorrect.
This can often refer to how much a trend is increasing or decreasing, such as \textit{rapidly} or \textit{slowly}, as well as whether it's a \textit{steady} change or \textit{highly-fluctuating} change.
In Appendix Figure \ref{app:qstability}, \textit{``The comparable sales growth of Sam's Club in the United States from fiscal year 2006 to 2020 has been \qualerr{steadily decreasing} from 2006 to 2020.''} should read \textit{``The comparable sales growth of Sam's Club in the United States from fiscal year 2006 to 2020 has been \qualfix{highly-fluctuatingly decreasing} from 2006 to 2020.''}
1.07\% (4) of the statements we analyzed contained this error.

\paragraph{Repetition.}
Repetition is when a caption repeats a previously-generated claim, regardless of its correctness.
117 (31.2\%) statements contained repetition, making it the most common error we encountered.
For example, in Appendix Figure \ref{app:qrepeat}, we see \textit{"The average age at widow hood in the Netherlands has increased from 2008 to 2018. The average age at widow hood in the Netherlands has increased from 2008 to 2018."}
Repetition is a known problem for text generation models with transformer backbones, like our chart captioning models~\citep{repetitionanalysis}.

\paragraph{Nonsensical Errors.}
If a L2 or L3 statement cannot be understood in context of the chart, or makes a fundamental mistake in interpretation, we label it as nonsensical error.
We encountered 20 nonsensical errors in addition to the 395 statements we analyzed.
For example, in Appendix Figure \ref{app:qnonsense}, \textit{"The \qualerr{most popular visitors was Harry Potter} in 1999 and 2009."} misinterprets the chart.
It might instead correctly read \textit{"The \qualfix{destination with the most visitors after the TV/movie's release} was New Zealand for The Lord of the Rings"}.
\section{Discussion}
\label{sec:discussion}
We present VisText, a chart captioning dataset of 12,441 charts and semantically rich captions.
The VisText charts are represented as a rasterized image, data table, and scene graph to provide diverse and complementary data modalities.
Using VisText, we fine-tune large language models to generate natural language captions from textual chart representations and integrate image-guided chart captioning to leverage multimodal information.
Utilizing the varied semantic content in VisText captions, we develop semantic prefix-tuned models that output semantically customized captions to meet diverse user needs.
Evaluations reveal that our models output precise and semantically descriptive captions, performing on par with state-of-the-art chart captioning models~\citep{kantharaj2022chart} across machine translation and text generation metrics.

Looking ahead, while accessibility remains a key domain that would benefit from automated chart captioning, and deploying automated chart captioning models into the field is an exciting prospect, we believe the most promising approach for future work lies in ``mixed-initiative'' (i.e., human + AI) chart authoring systems. 
In particular, as we describe in our Ethics Statement below, chart captioning models are currently prone to make a number of factual inaccuracies which can have severe harmful consequences.
On the other hand, by integrating these models into chart authoring systems (e.g., Tableau, Charticulator, Data Illustrator, or Lyra), chart authors can intervene and make any necessary corrections. 
Indeed, such integration offers exciting opportunities to develop novel interactive methods for verifying generated captions. 
For instance, models like ours could generate an initial caption (or set of captions) based on the chart currently being authored; as the system has access to all three representations of the chart (the backing data table, chart image, and structured scene graph), it might automatically segment the caption into independent ``data segments'' and interactively link and map them to rows in the table or regions on the chart, akin to Kori~\citep{latif2021kori}.

\section*{Limitations}
\label{sec:limits}

\paragraph{Computational Constraints.} Despite using modern GPUs, with large amounts of memory, we were forced to use the smallest-parameter variants of T5 and ByT5 as we encountered out-of-memory errors with the larger alternatives.
More problematically, the quadratic relationship between sequence length and time/space complexity of transformer architectures~\citep{attention}, especially when using byte-level sequences~\citep{byt5}, has had a significant impact on our model performance. 
In particular, to be computationally tractable, we were forced us to truncate our input and output sequences to, at most, 1,024 and 512 characters respectively (1,024 coming from the underlying ByT5 architecture~\citep{byt5}).

These character thresholds have likely had an outsized effect on \texttt{scene-graph} models.
For instance, due to these character limits, we reduced scene graph sequences to only a minimal set of visual characteristics; VisText also includes the raw, unprocessed scene graphs which offer a richer source of information about the visual features that are important to how people decode charts (e.g., bounding boxes, color) but were unavailable to our models.
Moreover, as Figure~\ref{fig:dataset_profile} shows, even with this reduced representation, the mean length of scene graph sequences is 948 characters (cf. 426 characters for data tables) with a wide distribution.
Thus, despite \texttt{scene-graph} models achieving comparable performance to \texttt{data-table} models, the former saw a much smaller proportion of complete sequences as compared to the latter.
This truncation step additionally negatively impacts charts with long titles or axis names\,---\,in such cases, we observed that the L2 or L3 caption would be altogether truncated before generation.

\paragraph{Chart Types and the Visualization Design Space.} 
VisText is scoped to only univariate bar, area, and line charts.
We chose to begin with these chart types informed by data visualization research that has focused on studying natural language descriptions of single-series line charts\,---\,a basic, but commonly occurring chart type that offers a compelling target of study as it most visibly surfaces any potential trends in the data~\citep{kim2021towards, stokes2022striking}. 
Future work can now begin to consider more complex chart forms in a step-by-step manner.
For instance, moving from univariate bar, area, and line charts to multivariate versions of these chart types (i.e., stacked bars and areas, grouped bars, and multi-series line charts). 
From there, work can also consider chart types that surface perceptual and cognitive phenomena in visually distinct ways (e.g., scatterplots, where trends appear as clusters of points; heatmaps, where color saturation often encodes a trend; or maps, where color or other layered elements such as symbols are used to represent data values).
Finally, automated methods for captioning visualizations may eschew chart typologies altogether in favor of visualization grammars\,---\,by offering a more composable and combinatorial approach to the design space~\citep{wilkinson2012grammar}, learning over visualization grammars may offer a more robust approach to captioning highly customized or unique visual forms. 

For each future work direction, we anticipate scene graph representations to prove more fruitful than the data table.
As the complexity of the visualization increases, its relationship to the data table only grows more ambiguous; the scene graph, on the other hand, directly encodes the visual form and thus remains faithful to it.
As a result, to support such future work, VisText provides the raw specifications used to produce our charts (via the Vega-Lite visualization grammar~\citep{satyanarayan2016vegalite}) as well as the raw, hierarchical scene graphs prior to our linearization and reduction step. 
\section*{Ethics Statement}
\label{sec:ethics}

\paragraph{The Consequences of Incorrect Captions.}
\citet{weidinger2021ethical} comprehensively survey the risks associated with the large language models (LLMs) that underlie our contribution. 
Of the six categories of risk they identify, harms stemming from models producing factually incorrect statements are not only most pertinent to our work, but are likely heighted as compared to general uses of LLMs given the context we are addressing: automatically captioning charts. 
In particular, people most often consume charts and visualizations in order to make data-driven decisions~\citep{keim2008visual}\,---\,for instance, about whether to evacuate ahead of a hurricane~\citep{padilla2018decision}, or health \& safety during the pandemic~\citep{shneiderman_data_2020}.
Moreover, recent results have shown that readers not only fixate for longer and are more likely to recall the textual content of and around visualizations~\citep{borkin2015beyond} but this textual content can strongly influence the takeaway message readers leave with even when it is at odds with the depicted data~\citep{kong2018frames,kong2019trust}.
Finally, these issues are exacerbated by the persuasive and rhetorical force of data and charts~\citep{kennedy2016work, hullman2011visualization}, that often project a sense of authority and certainty~\citep{correll2019ethical}. 
As a result, readers may not think to double check the accuracy of chart captions, and inaccurate statements that models may produce could lead to harmful downstream decisions. 

To proceed ethically with this line of research, we believe that advances in data and modeling need to be closely followed by attention devoted to mitigating the risks of incorrect statements. 
At base, automatically generated captions should be identified as such at the forefront to raise readers' awareness about the potential for incorrect statements.
And, interactive visual linking strategies (such as those explored by \citet{kong2012graphical, kim2018facilitating}) could be deployed to help readers manually verify the constituent statements of a caption against the chart. 
These strategies, however, place the burden of harm mitigation on readers.
Thus, an alternate approach might never surface automatically generated captions to readers directly but instead use them as part of mixed-initiative systems for jointly authoring visualization and text, such as Kori~\citep{latif2021kori}.
In such systems, automated chart captioning models would help to accelerate the authoring process\,---\,combatting the blank slate problem by providing an initial summary of the chart\,---\,and chart authors would make any necessary corrections prior to publication.

Besides these human-computer interaction (HCI) approaches for mitigating harm, an equally important direction for future work should leverage interpretability techniques to more deeply study what the models are learning.
To what degree are chart captioning models stochastic parrots~\citep{bender2021dangers}, and how much do they understand the information charts depict?

\paragraph{Automated Captioning for Accessibility.}
Although accessibility is a guiding motivation for the bulk of work in automated captioning (be it image captioning or, as in our case, chart captioning), studies find mixed reactions, at best, about these approaches among people with disabilities (PWDs).
For instance, accessibility educator and researcher Chancey Fleet described Facebook's automatic image descriptions as \textit{``famously useless in the Blind community''} despite \textit{``garner[ing] a ton of glowing reviews from mainstream outlets''}~\citep{fleet_things_2021, hanley2021computer}.
This disconnect appears to stem from a more fundamental mismatch between what PWDs describe as their captioning needs, and what the research community\,---\,particularly through its automatic, quantitative evaluations\,---\,prioritizes~\citep{jandrey2021image}. 
In particular, surveys with PWDs repeatedly surface the contextual nature of captions.
\citet{bennett2021s} find that the context of use shapes the degree to which PWD are comfortable with captions describing people's race, gender, and disabilities\,---\,for instance, changing their preferences if they were in a white, cisgender, nondisabled, and professional company versus their own community. 
Similarly, \citet{jung2022so} find shifting preferences for the content image descriptions should convey across different photo activites\,---\,for example, when viewing or taking photos, participants wished for descriptions that conveyed spatial cues whereas when searching or reminiscing about photos, participants hoped for descriptions to connect to personal data or differentiating details. 

In contrast, quantitative metrics of model performance compare generated captions to a single ``ground truth'' caption. 
This framing of success not only makes it difficult to develop contextually-varying caption generation but can actively penalize such investigations. 
For instance, with our work, we explored how prefix-tuning can be used to develop models that are responsive to users' preferences about semantic content. 
However, as described in Sec.~\ref{sec:quanteval}, existing quantitative metrics of model performance (e.g., BLEU, ROUGE, WMD, and TER) show a drop in model performance despite our qualitative analysis indicating that these captions are indeed high quality. 

Finally, our exploration of semantic prefix-tuning represents only a very preliminary step towards addressing the contextual captioning needs of PWDs. 
In particular, the semantic labels VisText assigns to captions were derived from prior work~\citep{lundgard2022accessible} that only explored natural language descriptions when \textit{consuming presentations} of visualizations\,---\,one task from a broader palette~\citep{brehmer2013multi}.
Future work might instead extend the VisText dataset\,---\,and corresponding models\,---\,to consider captions for a broader range of tasks including consuming visualizations for scientific discovery, enjoyment or, producing, searching, or querying visualizations~\citep{brehmer2013multi}.

\section*{Acknowledgements}

We thank Nicol\'{a}s Kennedy and Alan Lundgard for their work developing an initial version of our crowdsourced study protocol.
This research was sponsored by a Google Research Scholar Award, an NSF Award \#1900991, the MLA@CSAIL initiative, and by the United States Air Force Research Laboratory under Cooperative Agreement Number FA8750-19-2-1000. The views and conclusions contained in this document are those of the authors and should not be interpreted as representing the official policies, either expressed or implied, of the United States Air Force or the U.S. Government. The U.S. Government is authorized to reproduce and distribute reprints for Government purposes notwithstanding any copyright notation herein.

\bibliography{main}
\bibliographystyle{acl_natbib}

\clearpage
\onecolumn
\appendix
\section{Example Model Outputs}
\begin{figure*}[th]
  \centering
  \begin{subfigure}{\textwidth}
  \includegraphics[width=\textwidth]{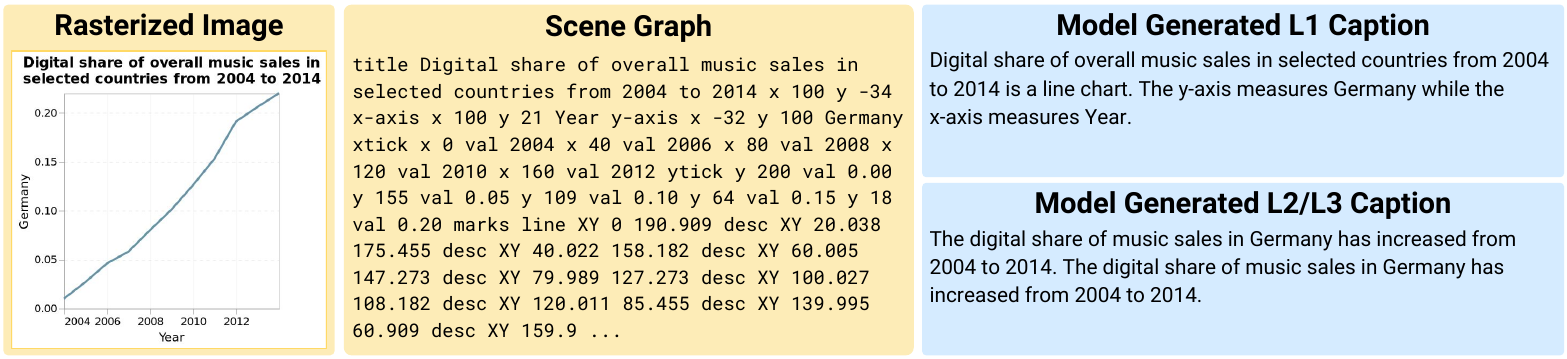}
  \caption{The \texttt{scene-graph} model's output L1 caption and L2/L3 caption for a VisText line chart of the \textit{``Digital share of overall music sales in selected counties from 2004 to 2014''}. The model correctly identifies the chart's title and axis, and it accurately identifies the upward trend. However, it repeats this claim twice. See Section~\ref{sec:qualeval} for details on model repetition.}
  \end{subfigure}
  
  \par\bigskip
  
  \begin{subfigure}{\textwidth}
  \includegraphics[width=\textwidth]{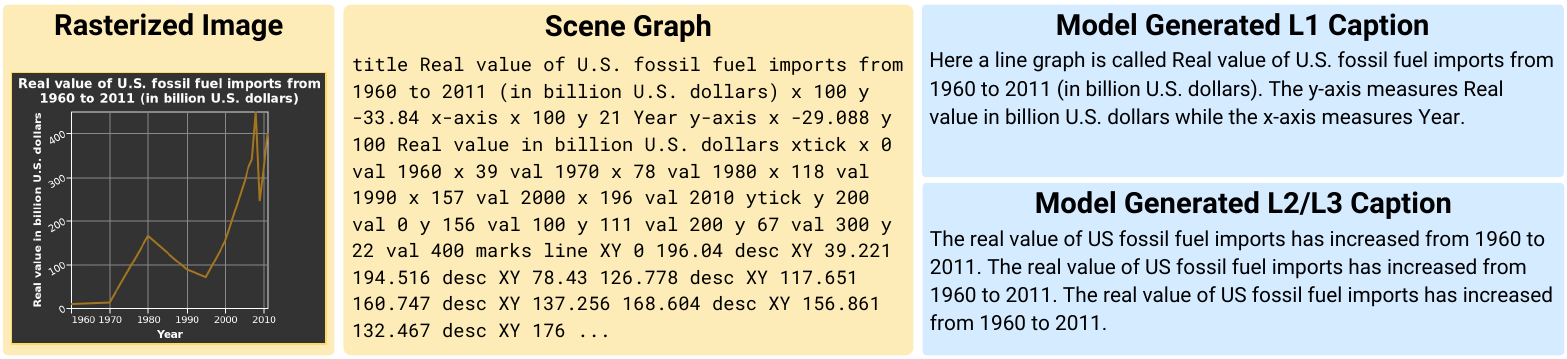}
  \caption{The \texttt{scene-graph} model's output L1 caption and L2/L3 caption for a VisText line chart of the \textit{``Real values of U.S. fossil fuel imports from 1960 to 2011 (in billion U.S. dollars)''}. The model correctly identifies the chart's title and axis, and it accurately identifies the upward trend. However, it repeats this claim twice. See Section~\ref{sec:qualeval} for details on model repetition.}
  \end{subfigure}
  
  \par\bigskip
  
  \begin{subfigure}{\textwidth}
  \includegraphics[width=\textwidth]{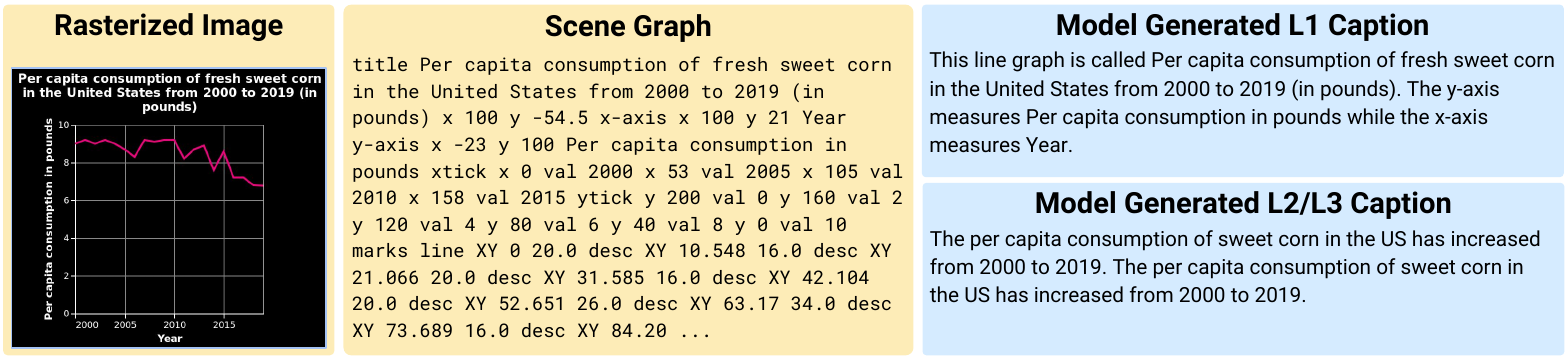}
  \caption{The \texttt{scene-graph} model's output L1 caption and L2/L3 caption for a VisText line chart of the \textit{``Per capita consumption of fresh sweet corn in the United States from 2000 to 2019 (in pounds)''}. The model correctly identifies the chart's title and axis. However, it makes a direction error and claims the increasing trend is actually decreasing. See Section~\ref{sec:qualeval} for details on direction errors.}
  \label{app:qdirection}
  \end{subfigure}
\caption{\texttt{scene-graph} model captions of VisText line charts.}
\end{figure*}
\begin{figure*}[th]
  \centering
  \begin{subfigure}{\textwidth}
  \includegraphics[width=\textwidth]{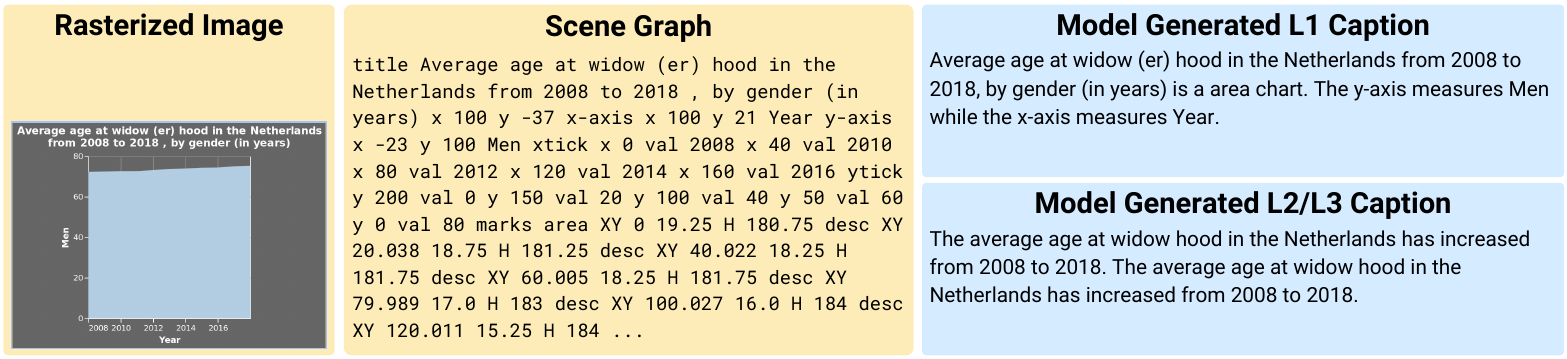}
  \caption{The \texttt{scene-graph} model's output L1 caption and L2/L3 caption for a VisText area chart of the \textit{``Average age at widow (er) hood in the Netherlands from 2008 to 2018, by gender (in years)''}. The model correctly identifies the chart's title and axis, and it accurately identifies the upward trend. However, it repeats this claim twice. See Section~\ref{sec:qualeval} for details on model repetition.}
  \label{app:qrepeat}
  \end{subfigure}
  
  \par\bigskip
  
  \begin{subfigure}{\textwidth}
  \includegraphics[width=\textwidth]{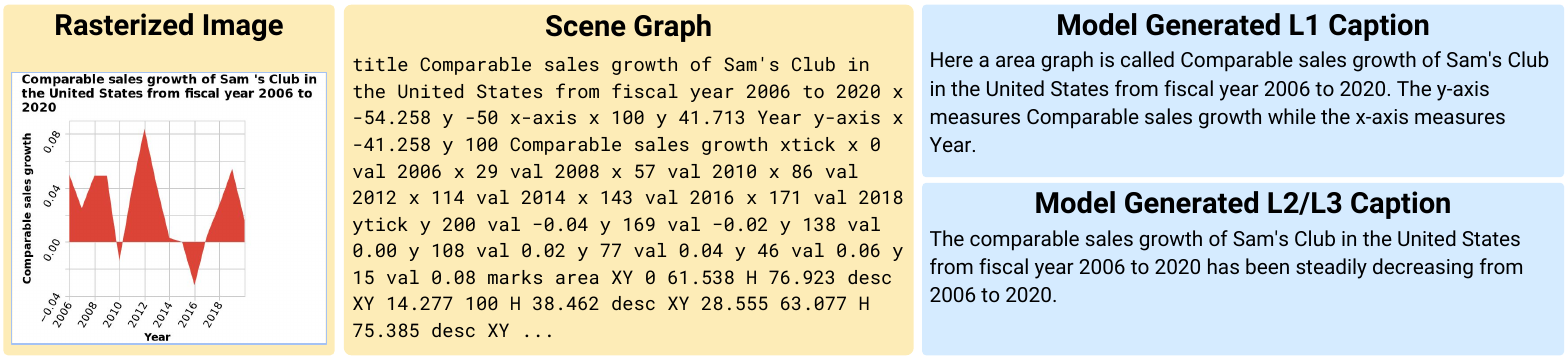}
  \caption{The \texttt{scene-graph} model's output L1 caption and L2/L3 caption for a VisText are chart of the \textit{``Comparable sales growth of Sam's Club in the United States from fiscal year 2006 to 2020''}. The model correctly identifies the chart's title and axis. However, it makes a stability error by stating the trend is steady when it is highly-fluctuating. See Section~\ref{sec:qualeval} for details on stability errors.}
  \label{app:qstability}
  \end{subfigure}
  
  \par\bigskip
  
  \begin{subfigure}{\textwidth}
  \includegraphics[width=\textwidth]{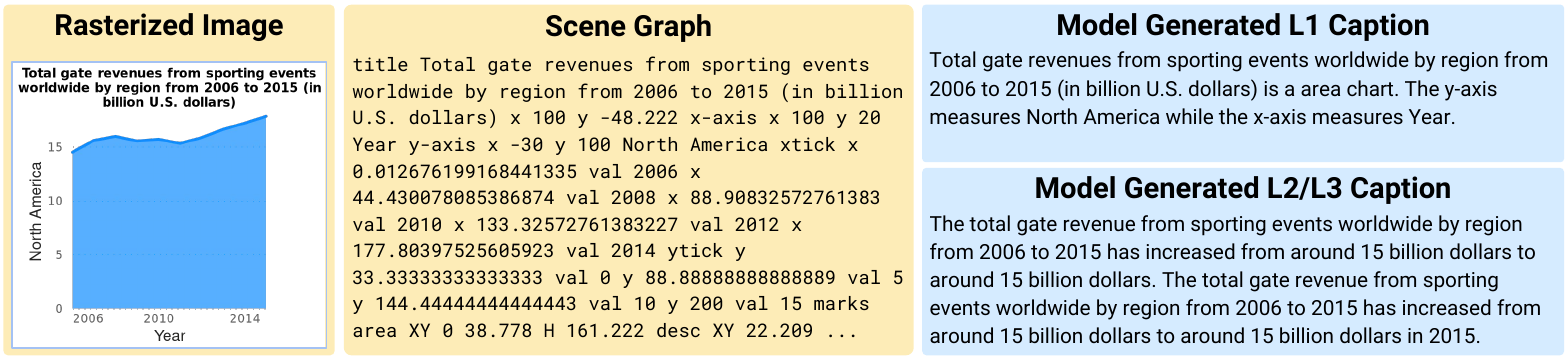}
  \caption{The \texttt{scene-graph} model's output L1 caption and L2/L3 caption for a VisText area chart of the \textit{``Total gate revenues from sporting events worldwide by region from 2006 to 2015 (in billion U.S. dollars)''}. The model correctly identifies the chart's title and axis. However, it makes a value error by claiming the revenue has increased to 15 billion dollars when it has actually increased to 18 billion dollars. See Section~\ref{sec:qualeval} for details on value errors.}
  \label{app:qvalue}
  \end{subfigure}
\caption{\texttt{scene-graph} model captions of VisText area charts.}
\end{figure*}
\begin{figure*}[th]
  \centering
  \begin{subfigure}{\textwidth}
  \includegraphics[width=\textwidth]{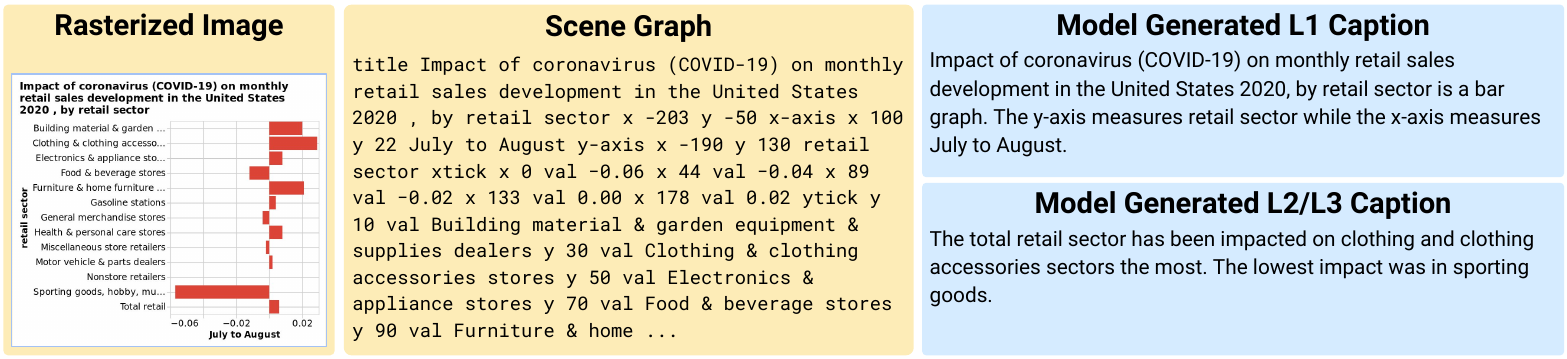}
  \caption{The \texttt{scene-graph} model's output L1 caption and L2/L3 caption for a VisText bar chart of the \textit{``Impact of coronavirus (COVID-19) on monthly retail sales development in the United States 2020, by retail sector''}. The model correctly identifies the chart's title and axis, and it correctly identifies the the most and least impacted sectors.}
  \end{subfigure}
  
  \par\bigskip
  
  \begin{subfigure}{\textwidth}
  \includegraphics[width=\textwidth]{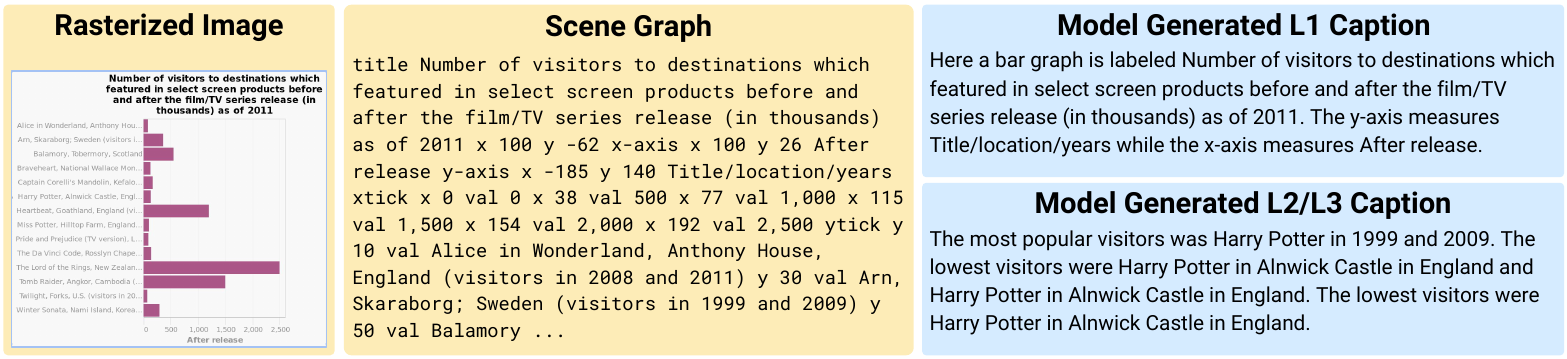}
  \caption{The \texttt{scene-graph} model's output L1 caption and L2/L3 caption for a VisText are chart of the \textit{``Number of visitors to destinations which featured in select screen products before and after the film/TV series release (in thousands) as of 2011''}. The model correctly identifies the chart's title and axis. However, it makes a nonsensical error by stating that most popular visitors was Harry Potter. See Section~\ref{sec:qualeval} for details on nonsensical errors.}
  \label{app:qnonsense}
  \end{subfigure}
  
  \par\bigskip
  
  \begin{subfigure}{\textwidth}
  \includegraphics[width=\textwidth]{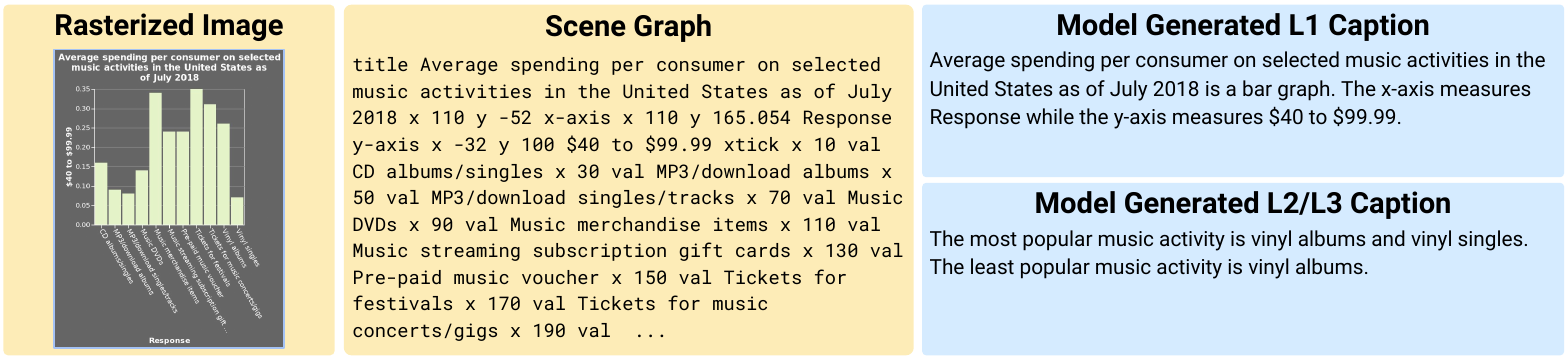}
  \caption{The \texttt{scene-graph} model's output L1 caption and L2/L3 caption for a VisText bar chart of the \textit{``Average spending per consumer on selected music activities in the United States as of July 2018)''}. The model correctly identifies the chart's title and axis. However, it makes an identity error by incorrectly identifying the most popular music activity. See Section~\ref{sec:qualeval} for details on identity errors.}
  \label{app:qidentity}
  \end{subfigure}
\caption{\texttt{scene-graph} model captions of VisText bar charts.}
\end{figure*}

\clearpage

\section{Additional Evaluations}

\subsection{Independent L1 and L2/L3 Caption Evaluation}
\begin{table*}

\begin{subtable}{\textwidth}
\centering
\resizebox{\linewidth}{!}{
\begin{tabular}{lrrrrrrrrrr}\toprule
\textbf{Input} &\textbf{Level} &\textbf{BLEU $\uparrow$} &\textbf{Perplexity $\downarrow$} &\textbf{ROUGE-1 $\uparrow$} &\textbf{ROUGE-2 $\uparrow$} &\textbf{ROUGE-L $\uparrow$} &\textbf{ROUGE-L SUM $\uparrow$} &\textbf{WMD $\downarrow$} &\textbf{TER $\downarrow$} \\\midrule
\citet{kantharaj2022chart} &L1 &\cellcolor[HTML]{a0cbe2} $0.43 \pm 1.33\mathrm{e}{-3}$ &\cellcolor[HTML]{e5f1f9} $67.58 \pm 2.56\mathrm{e}{+0}$ &\cellcolor[HTML]{a9d1e5} $0.73 \pm 3.00\mathrm{e}{-4}$ &\cellcolor[HTML]{acd2e6} $0.61 \pm 1.00\mathrm{e}{-3}$ &\cellcolor[HTML]{a5cee4} $0.63 \pm 1.33\mathrm{e}{-3}$ &\cellcolor[HTML]{a7cfe4} $0.63 \pm 1.40\mathrm{e}{-3}$ &\cellcolor[HTML]{b8d8e9} $0.58 \pm 2.67\mathrm{e}{-3}$ &\cellcolor[HTML]{a4cde3} $53.48 \pm 2.19\mathrm{e}{-2}$ \\ \midrule
\texttt{scene-graph} &L1 &\cellcolor[HTML]{9fcbe2} $0.43 \pm 4.67\mathrm{e}{-3}$ &\cellcolor[HTML]{bedbeb} $61.01 \pm 3.41\mathrm{e}{+0}$ &\cellcolor[HTML]{a3cde3} $0.74 \pm 2.70\mathrm{e}{-3}$ &\cellcolor[HTML]{a8d0e5} $0.61 \pm 6.80\mathrm{e}{-3}$ &\cellcolor[HTML]{9ecae1} $0.63 \pm 5.67\mathrm{e}{-4}$ &\cellcolor[HTML]{9ecae1} $0.63 \pm 5.33\mathrm{e}{-4}$ &\cellcolor[HTML]{a0cbe1} $0.57 \pm 1.72\mathrm{e}{-2}$ &\cellcolor[HTML]{9ecae1} $53.05 \pm 2.60\mathrm{e}{-1}$ \\
\texttt{data-table} &L1 &\cellcolor[HTML]{9fcbe2} $0.42 \pm 2.73\mathrm{e}{-3}$ &\cellcolor[HTML]{f7fbff} $70.68 \pm 4.33\mathrm{e}{+0}$ &\cellcolor[HTML]{a4cee3} $0.74 \pm 5.40\mathrm{e}{-3}$ &\cellcolor[HTML]{a5cee4} $0.62 \pm 6.50\mathrm{e}{-3}$ &\cellcolor[HTML]{a3cde3} $0.63 \pm 2.70\mathrm{e}{-3}$ &\cellcolor[HTML]{a3cde3} $0.63 \pm 2.97\mathrm{e}{-3}$ &\cellcolor[HTML]{a5cee3} $0.57 \pm 1.33\mathrm{e}{-2}$ &\cellcolor[HTML]{a1cbe2} $53.27 \pm 4.48\mathrm{e}{-1}$ \\ \midrule
\texttt{image} &L1 &\cellcolor[HTML]{f7fbff} $0.09 \pm 2.20\mathrm{e}{-3}$ &\cellcolor[HTML]{f5f9fe} $70.14 \pm 4.24\mathrm{e}{+0}$ &\cellcolor[HTML]{f7fbff} $0.40 \pm 2.50\mathrm{e}{-3}$ &\cellcolor[HTML]{f7fbff} $0.20 \pm 3.47\mathrm{e}{-3}$ &\cellcolor[HTML]{f7fbff} $0.30 \pm 2.50\mathrm{e}{-3}$ &\cellcolor[HTML]{f7fbff} $0.30 \pm 2.47\mathrm{e}{-3}$ &\cellcolor[HTML]{f7fbff} $1.09 \pm 8.33\mathrm{e}{-3}$ &\cellcolor[HTML]{f7fbff} $83.69 \pm 3.25\mathrm{e}{-1}$ \\
\texttt{image-scene-graph} &L1 &\cellcolor[HTML]{a1cce2} $0.43 \pm 2.35\mathrm{e}{-3}$ &\cellcolor[HTML]{9ecae1} $55.66 \pm 7.58\mathrm{e}{+0}$ &\cellcolor[HTML]{a0cbe2} $0.74 \pm 1.65\mathrm{e}{-3}$ &\cellcolor[HTML]{9ecae1} $0.62 \pm 1.10\mathrm{e}{-3}$ &\cellcolor[HTML]{a7cfe4} $0.63 \pm 1.50\mathrm{e}{-3}$ &\cellcolor[HTML]{a7cfe4} $0.63 \pm 1.50\mathrm{e}{-3}$ &\cellcolor[HTML]{b0d4e7} $0.58 \pm 6.05\mathrm{e}{-3}$ &\cellcolor[HTML]{a0cbe1} $53.22 \pm 1.27\mathrm{e}{-1}$ \\
\texttt{image-data-table} &L1 &\cellcolor[HTML]{a1cce2} $0.43 \pm 1.53\mathrm{e}{-3}$ &\cellcolor[HTML]{d3e7f3} $64.57 \pm 1.05\mathrm{e}{+0}$ &\cellcolor[HTML]{9ecae1} $0.74 \pm 1.07\mathrm{e}{-3}$ &\cellcolor[HTML]{a0cbe2} $0.62 \pm 1.43\mathrm{e}{-3}$ &\cellcolor[HTML]{abd1e6} $0.63 \pm 2.20\mathrm{e}{-3}$ &\cellcolor[HTML]{abd1e6} $0.63 \pm 2.27\mathrm{e}{-3}$ &\cellcolor[HTML]{a9d0e4} $0.57 \pm 4.47\mathrm{e}{-3}$ &\cellcolor[HTML]{a4cde3} $53.47 \pm 6.81\mathrm{e}{-2}$ \\
\bottomrule
\end{tabular}
}
\caption{Model results using the L1 captions.}
\label{tab:quant-L1}
\end{subtable}

\bigskip

\begin{subtable}{\textwidth}
\centering
\resizebox{\linewidth}{!}{
\begin{tabular}{lrrrrrrrrrr}\toprule
\textbf{Input} &\textbf{Level} &\textbf{BLEU $\uparrow$} &\textbf{Perplexity $\downarrow$} &\textbf{ROUGE-1 $\uparrow$} &\textbf{ROUGE-2 $\uparrow$} &\textbf{ROUGE-L $\uparrow$} &\textbf{ROUGE-L SUM $\uparrow$} &\textbf{WMD $\downarrow$} &\textbf{TER $\downarrow$} \\\midrule
\citet{kantharaj2022chart} &L2/L3 &\cellcolor[HTML]{b0d4e7} $0.07 \pm 7.67\mathrm{e}{-4}$ &\cellcolor[HTML]{f7fbff} $41.17 \pm 1.52\mathrm{e}{+0}$ &\cellcolor[HTML]{9ecae1} $0.30 \pm 2.57\mathrm{e}{-3}$ &\cellcolor[HTML]{9ecae1} $0.12 \pm 1.27\mathrm{e}{-3}$ &\cellcolor[HTML]{9ecae1} $0.26 \pm 1.80\mathrm{e}{-3}$ &\cellcolor[HTML]{9ecae1} $0.26 \pm 2.03\mathrm{e}{-3}$ &\cellcolor[HTML]{aad1e5} $0.92 \pm 6.00\mathrm{e}{-4}$ &\cellcolor[HTML]{9ecae1} $94.95 \pm 1.53\mathrm{e}{+0}$ \\ \midrule
\texttt{scene-graph} &L2/L3 &\cellcolor[HTML]{aed3e7} $0.07 \pm 8.07\mathrm{e}{-3}$ &\cellcolor[HTML]{dbebf5} $18.81 \pm 3.74\mathrm{e}{+0}$ &\cellcolor[HTML]{c3dfee} $0.28 \pm 1.65\mathrm{e}{-2}$ &\cellcolor[HTML]{cfe5f2} $0.11 \pm 9.43\mathrm{e}{-3}$ &\cellcolor[HTML]{b8d9ea} $0.25 \pm 1.02\mathrm{e}{-2}$ &\cellcolor[HTML]{bbdaeb} $0.24 \pm 1.02\mathrm{e}{-2}$ &\cellcolor[HTML]{acd1e5} $0.92 \pm 8.90\mathrm{e}{-3}$ &\cellcolor[HTML]{c8e1ef} $120.62 \pm 6.72\mathrm{e}{+0}$ \\
\texttt{data-table} &L2/L3 &\cellcolor[HTML]{a2cce3} $0.07 \pm 4.27\mathrm{e}{-3}$ &\cellcolor[HTML]{f7fbff} $23.90 \pm 2.75\mathrm{e}{+0}$ &\cellcolor[HTML]{add2e6} $0.30 \pm 1.09\mathrm{e}{-2}$ &\cellcolor[HTML]{bbdaeb} $0.11 \pm 6.50\mathrm{e}{-3}$ &\cellcolor[HTML]{a6cee4} $0.25 \pm 7.40\mathrm{e}{-3}$ &\cellcolor[HTML]{a6cee4} $0.25 \pm 7.43\mathrm{e}{-3}$ &\cellcolor[HTML]{a3cde2} $0.92 \pm 1.21\mathrm{e}{-2}$ &\cellcolor[HTML]{b9d9ea} $111.76 \pm 8.77\mathrm{e}{+0}$ \\ \midrule
\texttt{image} &L2/L3 &\cellcolor[HTML]{f7fbff} $0.02 \pm 2.73\mathrm{e}{-3}$ &\cellcolor[HTML]{9ecae1} $7.64 \pm 7.19\mathrm{e}{-1}$ &\cellcolor[HTML]{f7fbff} $0.17 \pm 1.22\mathrm{e}{-2}$ &\cellcolor[HTML]{f7fbff} $0.03 \pm 3.83\mathrm{e}{-3}$ &\cellcolor[HTML]{f7fbff} $0.14 \pm 1.02\mathrm{e}{-2}$ &\cellcolor[HTML]{f7fbff} $0.14 \pm 1.02\mathrm{e}{-2}$ &\cellcolor[HTML]{f7fbff} $1.19 \pm 7.40\mathrm{e}{-3}$ &\cellcolor[HTML]{f7fbff} $148.95 \pm 1.79\mathrm{e}{+1}$ \\
\texttt{image-scene-graph} &L2/L3 &\cellcolor[HTML]{c2deee} $0.06 \pm 4.85\mathrm{e}{-3}$ &\cellcolor[HTML]{dcecf6} $19.08 \pm 6.66\mathrm{e}{-1}$ &\cellcolor[HTML]{eff7fd} $0.26 \pm 1.13\mathrm{e}{-2}$ &\cellcolor[HTML]{e0eef7} $0.10 \pm 6.00\mathrm{e}{-3}$ &\cellcolor[HTML]{ecf5fc} $0.22 \pm 6.65\mathrm{e}{-3}$ &\cellcolor[HTML]{ecf5fc} $0.22 \pm 6.55\mathrm{e}{-3}$ &\cellcolor[HTML]{b5d7e8} $0.93 \pm 1.50\mathrm{e}{-3}$ &\cellcolor[HTML]{f7fbff} $151.28 \pm 1.17\mathrm{e}{+1}$ \\
\texttt{image-data-table} &L2/L3 &\cellcolor[HTML]{b8d9ea} $0.06 \pm 5.13\mathrm{e}{-3}$ &\cellcolor[HTML]{dcecf5} $19.02 \pm 1.79\mathrm{e}{+0}$ &\cellcolor[HTML]{d5e8f4} $0.27 \pm 5.00\mathrm{e}{-3}$ &\cellcolor[HTML]{c9e2f0} $0.11 \pm 3.23\mathrm{e}{-3}$ &\cellcolor[HTML]{d2e7f3} $0.23 \pm 3.47\mathrm{e}{-3}$ &\cellcolor[HTML]{d2e7f3} $0.23 \pm 3.53\mathrm{e}{-3}$ &\cellcolor[HTML]{a3cde2} $0.92 \pm 2.20\mathrm{e}{-3}$ &\cellcolor[HTML]{eff6fc} $144.20 \pm 6.11\mathrm{e}{+0}$ \\
\bottomrule
\end{tabular}
}
\caption{Model results using the L2/L3 captions.}
\label{tab:quant-L2L3}
\end{subtable}

\caption{We separately evaluate our L1 and L2L3 captions on all the same metrics except for Relation Generation. In general, we observe that L1 captions perform better than the L2/L3 captions. Our models generate verbose L1 captions that are similar to the structure of our L1 templates, while the L2/L3 captions are human-generated and contain more variability. Darker colors indicate better scores.}
\label{tab:quant-L1L2L3}
\end{table*}

To better understand how our models generate varying levels of semantic content, we separately evaluate our prefix-tuned models on L1 captioning and L2/L3 captioning tasks.
Each prefix-tuned model can output an L1 or an L2/L3 caption for each chart.
We evaluate these captions to their respective L1 or L2/L3 ground truth captions and report the results in Table~\ref{tab:quant-L1L2L3}.

Since we compute Relation Generation using only the L1 chart fields (e.g., chart title, axis scale, etc.), we do not report the results separately for L1 versus L2/L3 captioning.
There is no direct Relation Generation analog for L2/L3 captions, since they are human-generated and do not follow a specific template.
The Relation Generation for L1 captions is identical to the Relation Generation for L1/L2/L3 captions reported in Table~\ref{tab:quant-results-table}.

\subsection{Evaluation Details}
\paragraph{Quantitative Model Performance Metrics.}
We evaluate our models using NLP and machine translation metrics, including BLUE~\citep{bleu, bleu2}, Perplexity, Relation Generation~\citep{relationgeneration}, ROUGE~\citep{rouge}, Word Mover's Distance (WMD), and Translation Edit Rate (TER)~\citep{ter1, ter2}.
We implement Relation Generation per \citet{relationgeneration}, use the Gensim implementation of WMD, and use the Hugging Face implementation~\citep{huggingface} for the remaining metrics.
\begin{itemize}
    \item \textit{BLEU:} BLEU requires several gold standard references. In our evaluation setup, we use the test set caption as a single reference.
    
    \item \textit{Perplexity:} We use a pretrained GPT-2 Medium model to compute Perplexity.

    \item \textit{Relation Generation:} The fields we evaluate on are the chart title, axis names, and axis scales (if any).

    \item \textit{Translation Edit Rate (TER):} Edits consist of deletions, additions, and substitutions, as present in SacreBLEU.
\end{itemize}

\paragraph{Qualitative Caption Evaluation.}
To produce our qualitative evaluation results (Sec.~\ref{sec:qualeval}), we iteratively evaluated randomly sampled captions until there was no more marginal information about they types of errors to be gained from evaluating more captions.
For each L2/L3 caption, we assess the number of independent, mutually-exclusive L2 and L3 claims/statements that are being made.
In comparison to evaluating at a sentence-level, this allows us to take a more nuanced approach that isn't limited by where the model has generated a full-stop.
This approach allows us to more-accurately evaluate factual precision without overly-penalizing for a single mistake.

An example might take the form of "\textit{The lowest value is X} \textbf{(claim 1)}\textit{, the highest value is Y} \textbf{(claim 2)}\textit{, and the second highest is Z} \textbf{(claim 3)}\textit{. Overall, it is increasing over time} \textbf{(claim 4)}\textit{.}"
We observe that the first sentence is a compound sentence that consists of three independent clauses, each with a single factual L2 claim, while the second sentence is a single factual L3 claim.
Let us assume that \textbf{claim 1} was factually incorrect.
If we evaluate at a sentence-level, then the entire first sentence comprising of \textbf{claim 1}, \textbf{claim 2}, and \textbf{claim 3} would be incorrect.
However, by breaking this caption into independent, mutually-exclusive claims, we can more precisely calculate the factual precision of our text generation.

\clearpage

\section{Ablation Studies}
\begin{table*}
\begin{subtable}{\textwidth}
\centering
\resizebox{\linewidth}{!}{
\begin{tabular}{lrrrrrrrrrrrr}\toprule
\textbf{Experiment} &\textbf{Input} &\textbf{PT} &\textbf{BLEU $\uparrow$} &\textbf{Perplexity $\downarrow$} &\textbf{RG $\uparrow$} &\textbf{ROUGE-1 $\uparrow$} &\textbf{ROUGE-2 $\uparrow$} &\textbf{ROUGE-L $\uparrow$} &\textbf{ROUGE-L SUM $\uparrow$} &\textbf{WMD $\downarrow$} &\textbf{TER $\downarrow$} \\\midrule
\multirow{5}{*}{\shortstack[l]{Transformer\\Backbone}} &\texttt{BART-base} & &\cellcolor[HTML]{b1d5e8}$0.27 \pm 5.03\mathrm{e}{-3}$ &\cellcolor[HTML]{f7fbff}$43.06 \pm 5.76\mathrm{e}{+0}$ &\cellcolor[HTML]{a8d0e5}$1.69 \pm 9.80\mathrm{e}{-3}$ &\cellcolor[HTML]{a8d0e5}$0.59 \pm 9.67\mathrm{e}{-4}$ &\cellcolor[HTML]{a2cce3}$0.43 \pm 4.73\mathrm{e}{-3}$ &\cellcolor[HTML]{aed3e7}$0.48 \pm 2.07\mathrm{e}{-3}$ &\cellcolor[HTML]{aed3e7}$0.48 \pm 2.00\mathrm{e}{-3}$ &\cellcolor[HTML]{9ecae1}$0.65 \pm 1.47\mathrm{e}{-3}$ &\cellcolor[HTML]{9fcae1}$67.08 \pm 1.50\mathrm{e}{-1} $ \\
&\texttt{T5-small} & &\cellcolor[HTML]{a5cee4}$0.30 \pm 3.53\mathrm{e}{-3}$ &\cellcolor[HTML]{ecf5fb}$27.34 \pm 1.85\mathrm{e}{+0}$ &\cellcolor[HTML]{9fcbe2}$1.82 \pm 3.33\mathrm{e}{-4}$ &\cellcolor[HTML]{aed3e7}$0.58 \pm 1.27\mathrm{e}{-3}$ &\cellcolor[HTML]{a5cee4}$0.42 \pm 2.83\mathrm{e}{-3}$ &\cellcolor[HTML]{abd1e6}$0.49 \pm 1.37\mathrm{e}{-3}$ &\cellcolor[HTML]{abd1e6}$0.49 \pm 1.50\mathrm{e}{-3}$ &\cellcolor[HTML]{bbdaeb}$0.67 \pm 1.60\mathrm{e}{-3}$ &\cellcolor[HTML]{9fcae1}$67.03 \pm 1.82\mathrm{e}{-1} $ \\
&\texttt{T5-small} &\checkmark &\cellcolor[HTML]{a6cfe4}$0.30 \pm 5.40\mathrm{e}{-3}$ &\cellcolor[HTML]{f7fbff}$29.52 \pm 1.35\mathrm{e}{+0}$ &\cellcolor[HTML]{cee5f2}$1.15 \pm 7.70\mathrm{e}{-2}$ &\cellcolor[HTML]{f7fbff}$0.52 \pm 1.06\mathrm{e}{-2}$ &\cellcolor[HTML]{b8d8ea}$0.35 \pm 8.77\mathrm{e}{-3}$ &\cellcolor[HTML]{f7fbff}$0.43 \pm 7.83\mathrm{e}{-3}$ &\cellcolor[HTML]{f7fbff}$0.43 \pm 7.90\mathrm{e}{-3}$ &\cellcolor[HTML]{f7fbff}$0.71 \pm 1.22\mathrm{e}{-2}$ &\cellcolor[HTML]{d9eaf5}$76.44 \pm 1.28\mathrm{e}{+0} $ \\
&\texttt{Ours (ByT5-small)} & &\cellcolor[HTML]{a0cbe2}$0.32 \pm 4.07\mathrm{e}{-3}$ &\cellcolor[HTML]{c2dded}$20.96 \pm 3.09\mathrm{e}{+0}$ &\cellcolor[HTML]{9ecae1}$1.82 \pm 2.67\mathrm{e}{-4}$ &\cellcolor[HTML]{c9e2f0}$0.56 \pm 1.42\mathrm{e}{-2}$ &\cellcolor[HTML]{acd2e6}$0.39 \pm 1.62\mathrm{e}{-2}$ &\cellcolor[HTML]{c5dfee}$0.47 \pm 1.04\mathrm{e}{-2}$ &\cellcolor[HTML]{c5dfee}$0.47 \pm 1.04\mathrm{e}{-2}$ &\cellcolor[HTML]{cae2f0}$0.68 \pm 8.33\mathrm{e}{-3}$ &\cellcolor[HTML]{add2e6}$69.34 \pm 2.31\mathrm{e}{+0} $ \\
&\texttt{Ours (ByT5-small)} &\checkmark &\cellcolor[HTML]{9fcbe2}$0.32 \pm 2.13\mathrm{e}{-3}$ &\cellcolor[HTML]{bbdaeb}$20.02 \pm 2.25\mathrm{e}{+0}$ &\cellcolor[HTML]{a2cce3}$1.78 \pm 4.25\mathrm{e}{-2}$ &\cellcolor[HTML]{c9e2f0}$0.56 \pm 6.70\mathrm{e}{-3}$ &\cellcolor[HTML]{acd2e6}$0.39 \pm 6.23\mathrm{e}{-3}$ &\cellcolor[HTML]{c8e1ef}$0.47 \pm 6.37\mathrm{e}{-3}$ &\cellcolor[HTML]{c8e1ef}$0.47 \pm 6.40\mathrm{e}{-3}$ &\cellcolor[HTML]{cde4f1}$0.68 \pm 1.23\mathrm{e}{-2}$ &\cellcolor[HTML]{c1ddec}$72.55 \pm 1.75\mathrm{e}{+0} $ \\ \midrule
\multirow{2}{*}{L1 Generation} &\texttt{new-seed} &\checkmark &\cellcolor[HTML]{9ecae1}$0.32 \pm 4.47\mathrm{e}{-3}$ &\cellcolor[HTML]{c7e0ee}$21.77 \pm 1.63\mathrm{e}{+0}$ &\cellcolor[HTML]{9ecae1}$1.82 \pm 0.00\mathrm{e}{+0}$ &\cellcolor[HTML]{c1dded}$0.57 \pm 8.87\mathrm{e}{-3}$ &\cellcolor[HTML]{aad1e5}$0.40 \pm 8.30\mathrm{e}{-3}$ &\cellcolor[HTML]{bedcec}$0.47 \pm 7.73\mathrm{e}{-3}$ &\cellcolor[HTML]{bedcec}$0.47 \pm 7.67\mathrm{e}{-3}$ &\cellcolor[HTML]{c5dfee}$0.67 \pm 3.70\mathrm{e}{-3}$ &\cellcolor[HTML]{bbdaea}$71.61 \pm 2.95\mathrm{e}{+0} $ \\
&\texttt{original-seed} &\checkmark &\cellcolor[HTML]{9fcbe2}$0.32 \pm 2.13\mathrm{e}{-3}$ &\cellcolor[HTML]{bbdaeb}$20.02 \pm 2.25\mathrm{e}{+0}$ &\cellcolor[HTML]{a2cce3}$1.78 \pm 4.25\mathrm{e}{-2}$ &\cellcolor[HTML]{c9e2f0}$0.56 \pm 6.70\mathrm{e}{-3}$ &\cellcolor[HTML]{acd2e6}$0.39 \pm 6.23\mathrm{e}{-3}$ &\cellcolor[HTML]{c8e1ef}$0.47 \pm 6.37\mathrm{e}{-3}$ &\cellcolor[HTML]{c8e1ef}$0.47 \pm 6.40\mathrm{e}{-3}$ &\cellcolor[HTML]{cde4f1}$0.68 \pm 1.23\mathrm{e}{-2}$ &\cellcolor[HTML]{c1ddec}$72.55 \pm 1.75\mathrm{e}{+0} $ \\
\bottomrule
\end{tabular}
}
\caption{Ablation study results using the combined L1L2L3 captions.}
\end{subtable}

\bigskip
\begin{subtable}{\textwidth}
\centering
\resizebox{\linewidth}{!}{
\begin{tabular}{lrrrrrrrrrrr}\toprule
\textbf{Experiment} &\textbf{Input} &\textbf{Level} & \textbf{BLEU $\uparrow$} &\textbf{Perplexity $\downarrow$} &\textbf{ROUGE-1 $\uparrow$} &\textbf{ROUGE-2 $\uparrow$} &\textbf{ROUGE-L $\uparrow$} &\textbf{ROUGE-L SUM $\uparrow$} &\textbf{WMD $\downarrow$} &\textbf{TER $\downarrow$} \\\midrule
\multirow{2}{*}{\shortstack[l]{Transformer\\Backbone}} &\texttt{T5-small } &L1 &\cellcolor[HTML]{a5cee4}$0.42 \pm 7.87\mathrm{e}{-3} $&\cellcolor[HTML]{f7fbff}$73.01 \pm 5.20\mathrm{e}{+0} $&\cellcolor[HTML]{f7fbff}$0.64 \pm 1.64\mathrm{e}{-2} $&\cellcolor[HTML]{f7fbff}$0.52 \pm 1.44\mathrm{e}{-2} $&\cellcolor[HTML]{f7fbff}$0.56 \pm 1.10\mathrm{e}{-2} $&\cellcolor[HTML]{f7fbff}$0.56 \pm 1.09\mathrm{e}{-2} $&\cellcolor[HTML]{f7fbff}$0.65 \pm 1.06\mathrm{e}{-2} $&\cellcolor[HTML]{f7fbff}$62.76 \pm 1.10\mathrm{e}{+0}$ \\
&\texttt{Ours (ByT5-small)} &L1 &\cellcolor[HTML]{a0cbe2}$0.43 \pm 4.67\mathrm{e}{-3} $&\cellcolor[HTML]{bddbeb}$61.01 \pm 3.41\mathrm{e}{+0} $&\cellcolor[HTML]{a2cce3}$0.74 \pm 2.70\mathrm{e}{-3} $&\cellcolor[HTML]{a6cfe4}$0.61 \pm 6.80\mathrm{e}{-3} $&\cellcolor[HTML]{9ecae1}$0.63 \pm 5.67\mathrm{e}{-4} $&\cellcolor[HTML]{9ecae1}$0.63 \pm 5.33\mathrm{e}{-4} $&\cellcolor[HTML]{a0cbe1}$0.57 \pm 1.72\mathrm{e}{-2} $&\cellcolor[HTML]{9ecae1}$53.05 \pm 2.60\mathrm{e}{-1} $\\
\midrule
\multirow{2}{*}{L1 Generation} &\texttt{new-seed} &L1 &\cellcolor[HTML]{9ecae1}$0.43 \pm 1.50\mathrm{e}{-3} $&\cellcolor[HTML]{e9f3fa}$68.24 \pm 1.49\mathrm{e}{+1} $&\cellcolor[HTML]{a2cce3}$0.74 \pm 4.67\mathrm{e}{-4} $&\cellcolor[HTML]{a5cee4}$0.62 \pm 6.67\mathrm{e}{-5} $&\cellcolor[HTML]{a0cbe2}$0.63 \pm 2.97\mathrm{e}{-3} $&\cellcolor[HTML]{a0cbe2}$0.63 \pm 3.07\mathrm{e}{-3} $&\cellcolor[HTML]{9ecae1}$0.56 \pm 1.19\mathrm{e}{-2} $&\cellcolor[HTML]{9fcae1}$53.17 \pm 1.72\mathrm{e}{-1} $\\
&\texttt{original-seed} &L1 &\cellcolor[HTML]{a0cbe2}$0.43 \pm 4.67\mathrm{e}{-3} $&\cellcolor[HTML]{bddbeb}$61.01 \pm 3.41\mathrm{e}{+0} $&\cellcolor[HTML]{a2cce3}$0.74 \pm 2.70\mathrm{e}{-3} $&\cellcolor[HTML]{a6cfe4}$0.61 \pm 6.80\mathrm{e}{-3} $&\cellcolor[HTML]{9ecae1}$0.63 \pm 5.67\mathrm{e}{-4} $&\cellcolor[HTML]{9ecae1}$0.63 \pm 5.33\mathrm{e}{-4} $&\cellcolor[HTML]{a0cbe1}$0.57 \pm 1.72\mathrm{e}{-2} $&\cellcolor[HTML]{9ecae1}$53.05 \pm 2.60\mathrm{e}{-1} $\\
\bottomrule
\end{tabular}
}
\caption{Ablation study results using the L1 captions.}
\end{subtable}

\bigskip
\begin{subtable}{\textwidth}
\centering
\resizebox{\linewidth}{!}{
\begin{tabular}{lrrrrrrrrrrr}\toprule
\textbf{Experiment} &\textbf{Input} &\textbf{Level} &\textbf{BLEU $\uparrow$} &\textbf{Perplexity $\downarrow$} &\textbf{ROUGE-1 $\uparrow$} &\textbf{ROUGE-2 $\uparrow$} &\textbf{ROUGE-L $\uparrow$} &\textbf{ROUGE-L SUM $\uparrow$} &\textbf{WMD $\downarrow$} &\textbf{TER $\downarrow$} \\\midrule
\multirow{2}{*}{\shortstack[l]{Transformer\\Backbone}} &\texttt{T5-small} &L2/L3 &\cellcolor[HTML]{bcdbeb} $0.06 \pm 2.67\mathrm{e}{-3}$ &\cellcolor[HTML]{f7fbff} $35.81 \pm 4.13\mathrm{e}{+0}$ &\cellcolor[HTML]{f7fbff} $0.25 \pm 6.43\mathrm{e}{-3}$ &\cellcolor[HTML]{f7fbff} $0.09 \pm 3.43\mathrm{e}{-3}$ &\cellcolor[HTML]{f7fbff} $0.22 \pm 5.73\mathrm{e}{-3}$ &\cellcolor[HTML]{f7fbff} $0.22 \pm 5.60\mathrm{e}{-3}$ &\cellcolor[HTML]{f7fbff} $0.99 \pm 8.70\mathrm{e}{-3}$ &\cellcolor[HTML]{a1cce2} $113.33 \pm 2.94\mathrm{e}{+0}$ \\
&\texttt{Ours (ByT5-small)} &L2/L3 &\cellcolor[HTML]{b4d7e9} $0.07 \pm 8.07\mathrm{e}{-3}$ &\cellcolor[HTML]{ddecf6} $18.81 \pm 3.74\mathrm{e}{+0}$ &\cellcolor[HTML]{b9d9ea} $0.28 \pm 1.65\mathrm{e}{-2}$ &\cellcolor[HTML]{bbdaeb} $0.11 \pm 9.43\mathrm{e}{-3}$ &\cellcolor[HTML]{b2d5e8} $0.25 \pm 1.02\mathrm{e}{-2}$ &\cellcolor[HTML]{b5d7e9} $0.244 \pm 1.02\mathrm{e}{-2}$ &\cellcolor[HTML]{a9d0e4} $0.92 \pm 8.90\mathrm{e}{-3}$ &\cellcolor[HTML]{b3d5e8} $120.62 \pm 6.72\mathrm{e}{+0}$ \\
\midrule
\multirow{2}{*}{L1 Generation} &\texttt{new-seed} &L2/L3 &\cellcolor[HTML]{9ecae1} $0.08 \pm 5.93\mathrm{e}{-3}$ &\cellcolor[HTML]{e9f3fa} $20.96 \pm 2.71\mathrm{e}{+0}$ &\cellcolor[HTML]{abd1e6} $0.29 \pm 5.77\mathrm{e}{-3}$ &\cellcolor[HTML]{9ecae1} $0.11 \pm 2.33\mathrm{e}{-3}$ &\cellcolor[HTML]{a1cce2} $0.25 \pm 5.30\mathrm{e}{-3}$ &\cellcolor[HTML]{a1cce2} $0.25 \pm 5.27\mathrm{e}{-3}$ &\cellcolor[HTML]{9ecae1} $0.91 \pm 1.83\mathrm{e}{-3}$ &\cellcolor[HTML]{a9d0e4} $116.36 \pm 1.11\mathrm{e}{+1}$ \\
&\texttt{original-seed} &L2/L3 &\cellcolor[HTML]{b4d7e9} $0.07 \pm 8.07\mathrm{e}{-3}$ &\cellcolor[HTML]{ddecf6} $18.81 \pm 3.74\mathrm{e}{+0}$ &\cellcolor[HTML]{b9d9ea} $0.28 \pm 1.65\mathrm{e}{-2}$ &\cellcolor[HTML]{bbdaeb} $0.11 \pm 9.43\mathrm{e}{-3}$ &\cellcolor[HTML]{b2d5e8} $0.25 \pm 1.02\mathrm{e}{-2}$ &\cellcolor[HTML]{b5d7e9} $0.244 \pm 1.02\mathrm{e}{-2}$ &\cellcolor[HTML]{a9d0e4} $0.92 \pm 8.90\mathrm{e}{-3}$ &\cellcolor[HTML]{b3d5e8} $120.62 \pm 6.72\mathrm{e}{+0}$ \\
\bottomrule
\end{tabular}
}
\caption{Ablation study results using the L2/L3 captions.}
\end{subtable}

\caption{We perform two ablation studies to measure the impact of our model architectures and L1 caption generation. Our Transformer Backbone ablation study compares our ByT5-small backbone to T5-small with and without prefix-tuning (PT) and BART-base. Our L1 Generation ablation study analyzes our stochastic L1 caption generation pipeline with different random seeds. We evaluate each model using machine translation and text generation metrics: BLEU, Perplexity, Relation Generation (RG), ROUGE-1, ROUGE-2, ROUGE-L, ROUGE-L SUM, Word Mover's Distance (WMD), and Translational Error Rate (TER). Darker colors indicate better scores.}
\label{tab:ablation}
\end{table*}

To evaluate our modeling and dataset design choices, we run ablation studies measuring the impact of our transformer model backbones and stochastic data generation pipeline. 
We report the results in Table~\ref{tab:ablation}.

\paragraph{Transformer Backbone.} 
To understand the impact of our token-free, byte-to-byte architecture ByT5 model backbone, we explore other large language models.
Specifically, we compare our 300M parameter ByT5-small model~\citep{byt5} with a 60M parameter T5-small~\citep{T5} and 140M parameter BART-base model~\citep{bart}.
We also apply prefix-tuning to the ByT5 and T5 models.
We cannot apply prefix-uning to BART because BART does not support multi-task learning.
Quantitatively, using ByT5 does not appear to significantly improve upon T5.
However, we theorize that ByT5's token-free paradigm increases the input sequence length by compressing more input text into fewer input tokens.

\paragraph{L1 Caption Generation.}
Since we generate L1 captions stochastically, we evaluate whether our initial randomization impacted the model's results.
We compare generate a second set of L1 captions using a different random seed.
We see the results are nearly identical across all metrics, indicating our dataset captures a diverse set of L1 captions.

\clearpage

\clearpage

\section{Implementation Details}
\label{sup:trainingdetails}

\begin{table*}
\centering
    \resizebox{\linewidth}{!}{
\begin{tabular}{lllrrlrrrrrr}
    \toprule
    \textbf{Model} & \textbf{PT} & \textbf{Seeds} & \textbf{Epochs} & \textbf{Batch Size} & \textbf{Optim.}  & \textbf{Adam $\beta1$} & \textbf{Adam $\beta2$} & \textbf{Adam $\epsilon$} & \textbf{Weight Decay} & \textbf{LR} \\
    \midrule
    \citet{kantharaj2022chart} &  & 10, 20, 30, 40, 50 & 50 & 2 & AdamW & 0.9 & 0.999 & $1e-08$ & Linear & $5e-05$ \\
    \citet{kantharaj2022chart} & \checkmark & 10, 20, 30, 40, 50 & 50 & 3 & AdamW & 0.9 & 0.999 & $1e-08$ & Linear & $5e-05$\\
    \midrule
    \texttt{scene-graph} & \checkmark & 10, 20, 30, 40, 50 & 50 & 3 & AdamW & 0.9 & 0.999 & $1e-08$ & Linear & $5e-05$  \\
    \texttt{data-table} & \checkmark & 10, 20, 30, 40, 50 & 50 & 4 & AdamW & 0.9 & 0.999 & $1e-08$ & Linear & $5e-05$ \\
    \texttt{scene-graph} &  & 10, 20, 30, 40, 50 & 50 & 3 & AdamW & 0.9 & 0.999 & $1e-08$ & Linear & $5e-05$ \\
    \texttt{data-table} &  & 10, 20, 30, 40, 50 & 50 & 4 & AdamW & 0.9 & 0.999 & $1e-08$ & Linear & $5e-05$ \\
    \midrule
    \texttt{image-scene-graph} &  & 9555, 16710, 23578 & 50 & 4 & AdamW & 0.9 & 0.999 & $1e-06$ & 0.01 & $5e-05$ \\
    \texttt{image-scene-graph} & \checkmark & 1393, 16983, 23814 & 50 & 4 & AdamW & 0.9 & 0.999 & $1e-06$ & 0.01 & $5e-05$ \\
    \texttt{image-data-table} &  & 7504, 9586, 32579 & 50 & 4 & AdamW & 0.9 & 0.999 & $1e-06$ & 0.01 & $5e-05$ \\
    \texttt{image-data-table} & \checkmark & 4120, 7625, 19179 & 50 & 4 & AdamW & 0.9 & 0.999 & $1e-06$ & 0.01 & $5e-05$ \\
    \texttt{image} &  & 13423, 17963, 29028 & 50 & 32 & AdamW & 0.9 & 0.999 & $1e-06$ & 0.01 & $5e-05$  \\
    \texttt{image} & \checkmark & 4650, 7434, 15249 & 50 & 32 & AdamW & 0.9 & 0.999 & $1e-06$ & 0.01 & $5e-05$ \\
    \midrule
    \texttt{BART-base scene-graph} &  & 10, 20, 30, 40, 50 & 50 & 2 & AdamW & 0.9 & 0.999 & $1e-08$ & Linear & $5e-05$ \\
    \texttt{T5-small scene-graph} &  & 10, 20, 30, 40, 50 & 50 & 2 & AdamW & 0.9 & 0.999 & $1e-08$ & Linear & $5e-05$ \\
    \texttt{T5-small scene-graph} & \checkmark & 10, 20, 30, 40, 50 & 50 & 3 & AdamW & 0.9 & 0.999 & $1e-08$ & Linear & $5e-05$ \\
    \texttt{new-seed scene-graph} & \checkmark & 10, 20, 30, 40, 50 & 50 & 3 & AdamW & 0.9 & 0.999 & $1e-08$ & Linear & $5e-05$ \\
    \bottomrule
    \end{tabular}
}
\caption{A summary of notable hyperparameters we used to train the baseline, text-based, image-guided, and ablation study models. For all parameters and code, see: \url{https://github.com/mitvis/vistext}.}
\label{tab:hyperparams}
\end{table*}

Code to train and evaluate our text-based and image-guided models is available at \url{https://github.com/mitvis/vistext}.
Table~\ref{tab:hyperparams} summarizes our model training parameters.

\subsection{Text-Based Chart Captioning}
To train our text-based chart captioning models, we use the Huggingface implementation of ByT5~\citep{huggingface}. 
Due to hardware limitations, we use the ByT5-small model, which has 300M parameters. 
We fine-tune each model for 50 epochs, using Adam optimization with a learning rate of $5\mathrm{e}{-05}$.
To fit the input features into GPU memory, we truncate the input text (i.e., scene graph or data table) to 1024 tokens and the output caption to 512 tokens.
We select the best model epochs based on the validation loss of the validation set.
See Table~\ref{tab:hyperparams} or the VisText GitHub repository\footnote{\label{note:github}\url{https://github.com/mitvis/vistext}} for each model's full training details and hyperparameters.

We train each model three times with and without prefix-tuning and report the mean and standard deviation in Table~\ref{tab:quant-results-table}.
We train each model on four NVidia V100 GPUs with 32GB of memory connected by an NVLink2 network.
With prefix-tuning, training, evaluation, and inference took approximately 39 hours for the \texttt{scene-graph} model and 11 hours for the \texttt{data-table} models.
Without prefix-tuning, training, evaluation, and inference took approximately 78 hours for the \texttt{scene-graph} model and 22 hours for the \texttt{data-table} models.
We estimate that we trained each model between 30 to 45 times to achieve our final results.

\subsection{Image-Guided Chart Captioning}
Our image-guided chart captioning models extend the VLT5 model~\citep{cho2021unifying}, which is a multimodal extension of T5-base. 
We extract visual features from VisText's chart images using Bottom-Up Feature Extraction~\citep{anderson2017bottom} and 36 bounding boxes per image.
After feature extraction, we fine-tune VLT5 on the VisText dataset for 50 epochs following the default VLT5 training protocol\footnote{\url{https://github.com/j-min/VL-T5}}~\citep{cho2021unifying}.
To fit the input features into GPU memory, we truncate the input text (i.e., scene graph or data table) to 1024 tokens and the output caption to 512 tokens.
After 50 epochs, we select the epoch with the lowest validation loss as the best model.
See Table~\ref{tab:hyperparams} or the VisText GitHub repository\footnotemark[2] for each model's full training details and hyperparameters.

We train each model three times with and without prefix-tuning and report the mean and standard deviation in Table~\ref{tab:quant-results-table}.
We train each model on four NVidia V100 GPUs with 1TB of memory.
The \texttt{image} models take approximately 2 minutes per training epoch without prefix-tuning and approximately 3 minutes per training epoch with prefix-tuning.
The \texttt{image-scene-graph} and \texttt{image-data-table} models take approximately 10 minutes per training epoch without prefix-tuning and approximately 15 minutes per training epoch with prefix-tuning.
We estimate that we trained each model between 5 to 10 times to achieved our final results.

\subsection{Ablation Models}
We train our ablation models using the same parameters as our default models, only varying the parameter of interest.
We train them on 16 virtual CPU cores on Xeon E5 hypervisors with 128GB of memory and PCI pass-through access to eight NVidia Titan XP GPUs with 12GB of memory.

\subsection{Notable Package Versions}
Package versions are listed in Table~\ref{tab:packages}.
\begin{table}
  \footnotesize
  \centering
  \begin{tabular}{lr}
  \toprule
  \textbf{Package} & \textbf{Version} \\
  \midrule
  bleurt &                    0.0.2 \\
  datasets &                  2.10.1 \\
  evaluate &                  0.4.0 \\
  gensim &                    4.3.0 \\
  h5py &                      3.7.0 \\
  nltk &                      3.7 \\
  numpy &                     1.22.3 \\
  pandas &                    1.4.2 \\
  pot &                       0.9.0 \\
  pytorch &                   1.10.2 \\
  pyyaml &                    5.4.1 \\
  sacrebleu &                 2.0.0 \\
  scipy &                     1.7.3 \\
  sentencepiece &             0.1.95 \\
  tokenizers &                0.11.4 \\
  torchvision &               0.13.1 \\
  transformers &              4.24.0 \\
  \bottomrule
  \end{tabular}

\caption{The package versions used for training and evaluating our VisText models. Further implementation details and code are available at: \url{https://github.com/mitvis/vistext}.}
\label{tab:packages}
\end{table}

\clearpage

\section{Additional VisText Dataset Details}
\label{sec:appdatasetgen}

\subsection{Licensing}
Our use of the raw Statista data from \citet{kantharaj2022chart} is consistent with its intended use case.
The data was licensed under the GNU General Public License v3.0.
We release our data and code under GNU General Public License v3.0.

\subsection{L1 Caption Generation Process}
\label{sec:l1_generation}

The Level 1 captions are generated from a random process that chooses from 3 title templates and 6 axis templates.
The title templates we use are:
\begin{itemize}
    \item This is a \texttt{[chart-type]} titled \texttt{[chart-title]}
    \item This \texttt{[chart-type]} is titled \texttt{[chart-title]}
    \item \texttt{[chart-title]} is a \texttt{[chart-type]}
\end{itemize}

\noindent The axis templates we use for each axis are:
\begin{itemize}
    \item On the \texttt{[axis]}, \texttt{[axis-label]} is plotted with a \texttt{[axis-scale]}
    \item \texttt{[axis-label]} is plotted with a \texttt{[axis-scale]} on the \texttt{[axis]}
    \item The \texttt{[axis]} plots \texttt{[axis-label]} with a \texttt{[axis-scale]}
    \item A \texttt{[axis-scale]} can be found on the \texttt{[axis]}, labeled \texttt{[axis-label]}
    \item This is a \texttt{[axis-scale]} on the \texttt{[axis]}, labeled \texttt{[axis-label]}
\end{itemize}

\noindent We additionally have one template for both axes:
\begin{itemize}
    \item The \texttt{[axis1]} plots \texttt{[axis1-label]} with \texttt{[axis1-scale]} while the \texttt{[axis2]} plots \texttt{[axis2-label]} with \texttt{[axis2-scale]}
\end{itemize}

\noindent For each axis template, we randomly choose whether to include the axis scale.
Furthermore, within each template, we further randomly swap words with synonyms. A list of words and their possible synonym substitutions are:
\begin{itemize}
    \item \textbf{this:} \textit{here, a}
    \item \textbf{chart:} \textit{graph, diagram, plot}
    \item \textbf{titled:} \textit{called, named, labeled}
    \item \textbf{on:} \textit{along}
    \item \textbf{plotted}: \textit{defined, measured, drawn, shown}
    \item \textbf{plots:} \textit{measures, shows}
    \item \textbf{with:} \textit{using, on, along, as}
    \item \textbf{found:} \textit{seen}
    \item \textbf{labeled:} \textit{marked}
\end{itemize}

\subsection{Crowdsourced Study Protocol}
\label{sec:crowdsource_materials}

\begin{figure*}[th]
  \centering
  \includegraphics[width=\textwidth]{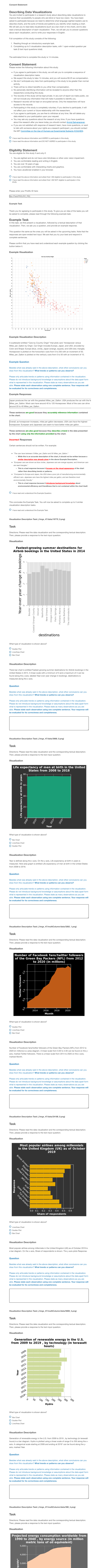}
    \caption{Introduction to the crowdsourcing study.}
  \label{fig:study_intro}
\end{figure*}

\begin{figure*}[th]
  \centering
  \includegraphics[width=\textwidth]{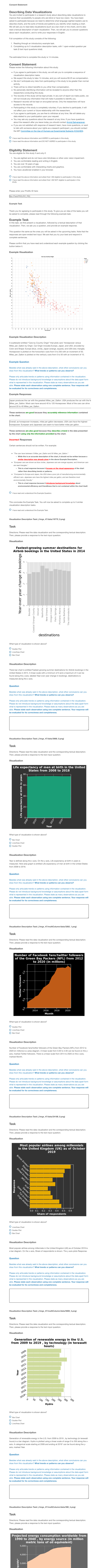}
    \caption{Eligibility statement for the crowdsourcing study.}
  \label{fig:study_eligibility}
\end{figure*}

\begin{figure*}[th]
  \centering
  \includegraphics[width=\textwidth]{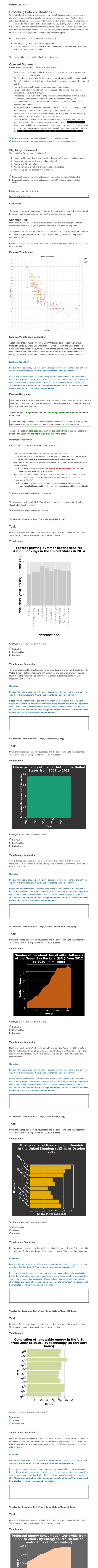}
    \caption{Consent statement for the crowdsourcing study.}
  \label{fig:study_consent}
\end{figure*}

\begin{figure*}[th]
  \centering
  \includegraphics[width=0.49\textwidth]{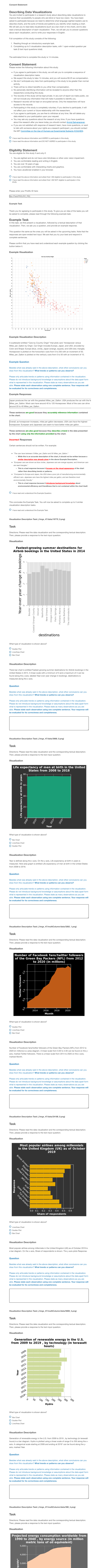}
    \caption{Instructions for the crowdsourcing study, with examples of correct and incorrect responses.}
  \label{fig:study_instructions}
\end{figure*}

\begin{figure*}[th]
  \centering
  \includegraphics[width=0.78\textwidth]{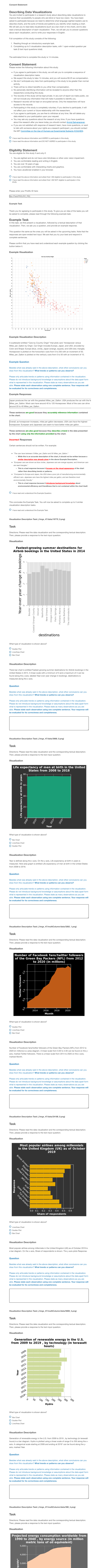}
    \caption{An example task from a specific run of the crowdsourcing study.}
  \label{fig:study_task}
\end{figure*}

Figures~\ref{fig:study_intro}--\ref{fig:study_task} screenshot the introduction, eligibility and consent statements, instructions, and a task from our crowdsourced study. 
We recruited participants on the Prolific.co crowdsourcing platform, following conventions in the data visualization research community\footnote{\url{https://twitter.com/eagereyes/status/1187773534745088000}} and recent research results~\citep{tang2022replication} that suggest Prolific yields higher quality results than Amazon Mechanical Turk.
We conducted multiple pilot runs to calibrate the amount of time it would take participants to complete the study, and found that most participants were able to successfully do so within 14 minutes. 
Following \citet{silberman2018responsible}, who advocate for paying workers at least minimum wage \textit{at your location}, we choose to pay our participants \$3.25\,---\,a roughly \$14/hour rate in line with the \$14.25/hour minimum wage in Massachusetts at the time the study was conducted. 

Our study was determined to be exempt by MIT's institutional review board (IRB).
Participants had to explicitly provide their consent in order to proceed with the study\,---\,if participants did not consent, they were redirected back to the Prolific platform.
The consent statement (Fig.~\ref{fig:study_consent}) reminded participants of their rights (including that their participation is voluntary and consent could be revoked at any time), and encouraged participants to contact either the study PI or IRB board directly should they have any concerns. 
We constrained our participant pool (and eligibility requirements) to people living within the United States or United Kingdom who self-reported as being sighted with no vision or color impairments. 
We did not collect any additional demographic data from participants as we did not determine this to bias or otherwise affect the content we hoped to collect. 

Each task (an example of which is shown in Fig.~\ref{fig:study_task}) included an attention check where participants were asked to correctly identify the chart type shown. 
If participants failed more than two attention checks, their submission was flagged for manual review\,---\,in practice, the bulk of participants who failed attention checks nevertheless produced valid captions and, thus, were paid fully.
The task asked participants to complete a free response question to describe as completely as they could the trends and patterns observed, emphasizing that their response would be evaluated for correctness and completeness.
Despite best practices suggesting a more structured, querying approach (called QID) can yield higher quality captions~\citep{morash2015guiding}, we opted for our free-response approach as the benefits of QID (namely, in expressing the chart type, title, and axes units) would already be captured by our synthetically generated L1 captions. 
Moreover, in contrast to the templatized output produced by QID, we hoped that our free-response responses would yield more ``natural'' articulations of perceptual and cognitive trends, following the \citet{lundgard2022accessible} framework.

\end{document}